%% file: main.tex
\documentclass[10pt,twocolumn,letterpaper]{article}

\usepackage[pagenumbers]{cvpr} 

\usepackage{booktabs}

\input{preamble}
\definecolor{cvprblue}{rgb}{0.21,0.49,0.74}
\usepackage[pagebackref,breaklinks,colorlinks,allcolors=cvprblue]{hyperref}

\def\paperID{45880} 
\def\confName{CVPR}
\def\confYear{2026}

\title{Single-step Diffusion-based Video Coding with Semantic-Temporal Guidance}

\author{
Naifu Xue$^{1*}$ \quad 
Zhaoyang Jia$^{2*}$ \quad 
Jiahao Li$^3$ \quad
Bin Li$^3$ \quad
Zihan Zheng$^2$\thanks{\scriptsize Naifu Xue, Zhaoyang Jia and Zihan Zheng are visiting students of MSRA.} \quad
Yuan Zhang$^1$ \quad
Yan Lu$^3$  \\
$^1$ Communication University of China \quad 
$^2$ University of Science and Technology of China\\
$^3$ Microsoft Research Asia\\
{\tt\small \{aaronxuenf, yzhang\}@cuc.edu.cn, \{jzy$\_$ustc, zzh2003\}@mail.ustc.edu.cn }\\
{\tt\small \{li.jiahao, libin, yanlu\}@microsoft.com}
}
\vspace{-10mm}

\begin{document}
\maketitle
\input{sec_camera/0_abstract}

\vspace{-4mm}
\input{sec_camera/1_intro}
\input{sec_camera/2_related}

\input{sec_camera/3_method}
\input{sec_camera/4_experiment}
{
    \small
    \bibliographystyle{ieeenat_fullname}
    \bibliography{main}
}

\clearpage
\maketitlesupplementary
\appendix
\input{sec_camera/5_suppl}

\end{document}

%% file: sec_camera/0_abstract.tex
\begin{abstract}
While traditional and neural video codecs (NVCs) have achieved remarkable rate–distortion performance, improving perceptual quality at low bitrates remains challenging.
Some NVCs incorporate perceptual or adversarial objectives but still suffer from artifacts due to limited generation capacity, whereas others leverage pretrained diffusion models to improve quality at the cost of high sampling complexity.
To overcome these challenges, we propose \textbf{S$^2$VC}, a \textbf{Single}-\textbf{S}tep diffusion–based \textbf{V}ideo \textbf{C}odec that integrates a conditional coding framework with an efficient single-step diffusion generator, enabling realistic reconstruction at low bitrates with reduced sampling cost.
Recognizing the importance of semantic conditioning in single-step diffusion, we introduce Contextual Semantic Guidance to extract frame-adaptive semantics from buffered features.
This guidance replaces text captions with efficient, fine-grained conditioning, thereby improving generation realism.
In addition, Temporal Consistency Guidance is incorporated into the diffusion U-Net to enforce temporal coherence across frames and ensure stable generation.
Extensive experiments show that S$^2$VC delivers state-of-the-art perceptual quality with an average bitrate saving of 51.62\% over prior perceptual method, underscoring the promise of single-step diffusion for efficient, high-quality video compression. 
Project: \url{https://onedc-codec.github.io/s2vc/}
\end{abstract}

%% file: sec_camera/1_intro.tex
\section{Introduction}
\label{sec:intro}

The exponential growth of video content has placed unprecedented demands on data storage and network bandwidth~\cite{cisco2020internet}, highlighting the importance of compact yet high-quality video compression.
Although recent traditional~\cite{bross2021overview, jvet2025ecm} and neural video codecs (NVCs)~\cite{li2023neural,DCVC-HEM,qi2024long, li2024neural, jia2025towards} have shown substantial improvements in rate–distortion (RD) performance, preserving perceptual quality at low bitrates remains challenging.
Traditional codecs (e.g., VVC~\cite{bross2021overview}) often suffer from visible blocking and ringing artifacts due to coarse quantization and block-wise prediction–transform design, whereas neural codecs~\cite{li2023neural, li2024neural, jia2025towards} optimized for objective distortions (MSE or MS-SSIM~\cite{wang2003multiscale}) tend to produce blurred frames. 
These limitations motivate research on perceptually oriented video codecs.

\begin{figure}[t]
    \centering
    \includegraphics[width=1.0\linewidth]{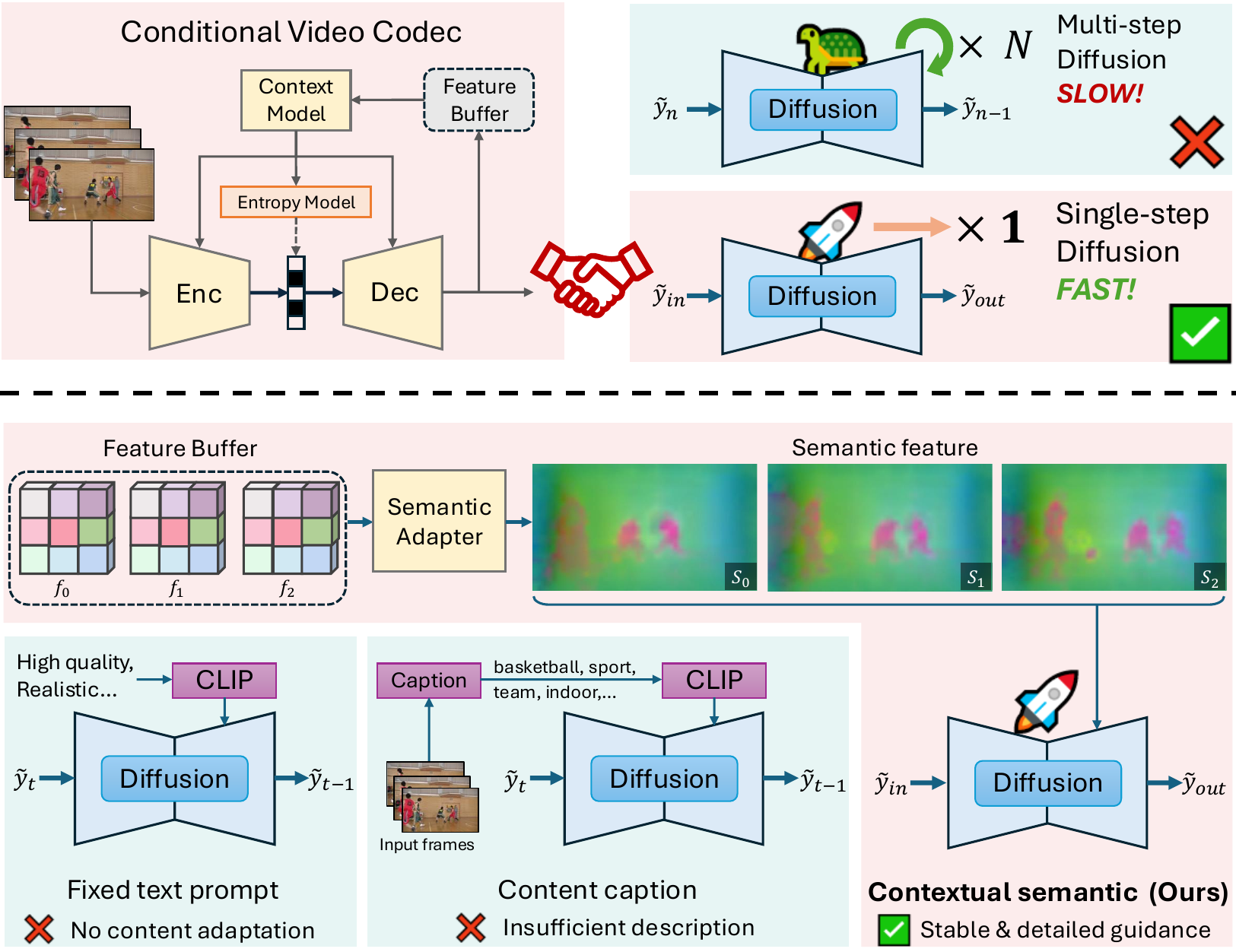}
    \vspace{-6mm}
    \caption{
        \textit{Top}: Our conditional video codec adopts a single-step diffusion model, which is especially critical for video, where multi-step diffusion would make sampling many frames prohibitively expensive.
        \textit{Bottom}: Comparison of semantic guidance. Fixed text prompts cannot adapt to dynamic video content, while captions lack fine-grained details. Our contextual semantic guidance provides detailed frame-wise information without requiring additional caption or embedding models.
    }
    \label{fig:1}
    \vspace{-6mm}
\end{figure}

\begin{figure*}[t]
    \centering
    \includegraphics[width=1.0\linewidth]{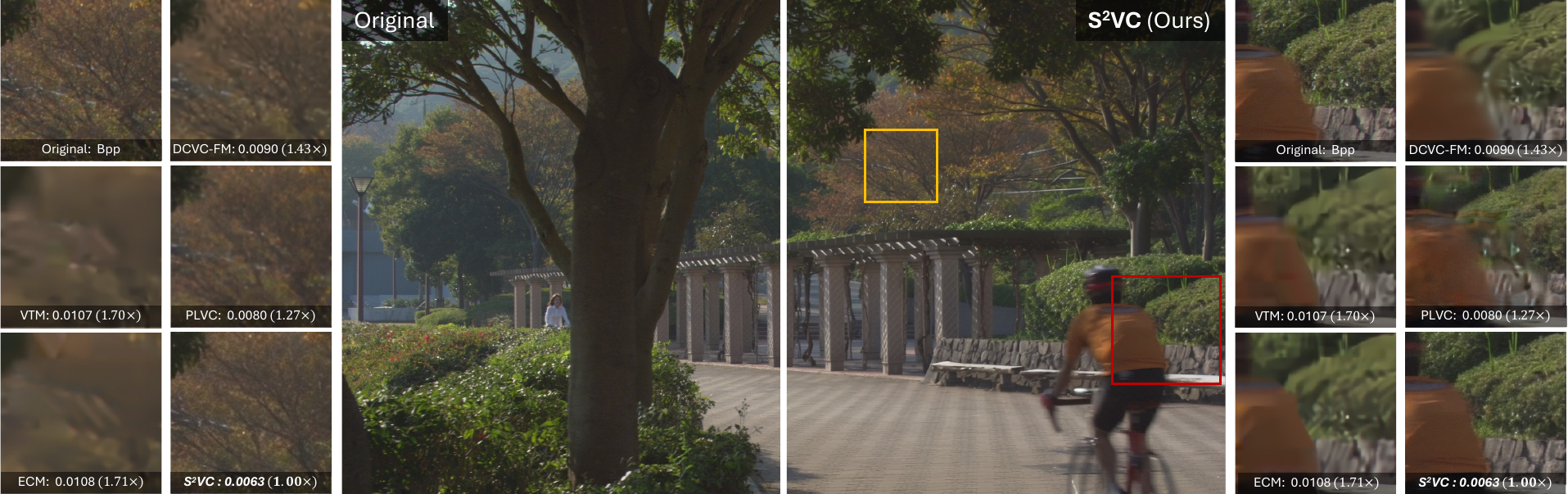}
    \vspace{-6mm}
    \caption{
        Example of decoded frames. 
        S$^2$VC delivers the best perceptual quality while maintaining the lowest bitrate.
        In contrast, traditional codecs~\cite{bross2021overview, jvet2025ecm} show blocking phenomenon, DCVC-FM~\cite{li2024neural} blurs details, and PLVC~\cite{yang2022perceptual} introduces artifacts.
    }
    \label{fig:2}
    \vspace{-5mm}
\end{figure*}

To improve perceptual quality, several studies~\cite{yang2022perceptual, li2023high} incorporate perceptual objectives and generative adversarial networks (GANs) into neural video compression, following the paradigm of generative image codecs~\cite{mentzer2020high, muckley2023improving}.
However, due to limited model capacity and training scale, these codecs still exhibit noticeable artifacts at low bitrates, compromising overall visual quality~\cite{ma2025diffusion}.
More recently, diffusion-based video codecs~\cite{liu20242, ma2025diffusion, ma2025diffvc} have leveraged pretrained image diffusion models~\cite{rombach2021highresolution} as frame reconstruction modules for perceptual enhancement.
While diffusion models demonstrate remarkable detail restoration, the resulting codecs still operate at relatively high bitrates, where traditional and neural-based methods already achieve visually acceptable quality.
In addition, the multi-step sampling process introduces substantial computational overhead~\cite{liu20242, ma2025diffusion}, even for a single image, further limiting their practicality in multi-frame video compression.
Therefore, reducing both bitrate and sampling steps is crucial for making diffusion-based video codecs more efficient.

To overcome these limitations, we draw inspiration from the success of single-step image diffusion models~\cite{yin2024one, yin2024improved} and propose \textbf{S$^2$VC}, a \textbf{Single}-\textbf{S}tep diffusion–based \textbf{V}ideo \textbf{C}odec designed for high-fidelity and temporally coherent reconstruction.
For codec architecture, S$^2$VC integrates an efficient single-step diffusion generator into a conditional coding framework~\cite{jia2025towards}, targeting low bitrate compression.
This design leverages large-scale generative priors to enhance perceptual quality while significantly reducing computational cost through single-step sampling, as illustrated in Fig.~\ref{fig:1} \textit{(top)}.
Moreover, the single-step formulation allows direct pixel-domain optimization, further improving reconstruction fidelity.
For generation enhancement, S$^2$VC introduces two complementary guidance mechanisms, \textit{Contextual Semantic Guidance (CSG)} and \textit{Temporal Consistency Guidance (TCG)}, which adapt the single-step image diffusion model for video decoding, achieving semantically accurate and temporally stable reconstruction.

Prior works~\cite{careil2023towards, li2024towards} often rely on fixed text prompts or generated captions as the semantic guidance for diffusion models.
Yet, the former lacks content adaptability, while the latter has difficulty conveying fine-grained details (see Fig.~\ref{fig:1}~\textit{bottom}).
To overcome these limitations, we propose \textit{Contextual Semantic Guidance (CSG)} as an improvement, especially considering the critical role of semantic guidance revealed in the prior single-step diffusion codec~\cite{xue2025one}.
Concretely, CSG extracts compact, frame-adaptive semantic representations from buffered features, replacing textual embeddings in the diffusion model.
This design provides accurate, content-aware guidance without heavy captioning or embedding networks, thereby improving generation realism while avoiding additional complexity.
To ensure the extracted semantics remain effective and temporally stable, we adopt DINOv3~\cite{simeoni2025dinov3} as a teacher for semantic distillation, leveraging its strong feature consistency in video scenarios.
Ablation results demonstrate that CSG substantially enhances final reconstruction quality.

Another critical aspect of generative video coding is temporal consistency.
The synthesized details should remain stable across consecutive frames for the same object; otherwise, flickering or jitter artifacts may arise, degrading perceptual quality.
To this end, we introduce \textit{Temporal Consistency Guidance (TCG)}, which guides the diffusion U-Net to generate temporally coherent textures.
Specifically, TCG inserts multiple blocks into the U-Net at different spatial scales.
Each block takes latent features from both the previous and current frames to extract temporal correlations, which are then transformed into diffusion guidance signals.
In cooperation with cascade training~\cite{li2023neural} that jointly optimizes quality across frames, the TCG effectively enhances temporal consistency, as validated by our ablation results.

Extensive experiments demonstrate that the proposed S$^2$VC achieves superior performance in both perceptual metrics and visual quality.
Notably, it attains state-of-the-art (SOTA) results across benchmarks, especially at bitrates below 0.02 bits per pixel (bpp).
As shown in Fig.~\ref{fig:2}, the reconstructed frames exhibit high perceptual fidelity and strong visual realism, further validating the effectiveness of S$^2$VC.
For quantitative evaluation, S$^2$VC achieves an average 51.62\% bitrate saving in frame-wise DISTS~\cite{ding2020image} compared with the previous perceptual codec~\cite{yang2022perceptual}.
It also attains strong performance on motion-aware FloLPIPS~\cite{danier2022flolpips} and realism-oriented FID~\cite{heusel2017gans}, underscoring the potential of single-step diffusion for generative video compression.

Our contributions are summarized as follows:
\begin{itemize}
    \item We propose S$^2$VC, a video codec that integrates a conditional coding pipeline with a single-step diffusion model, targeting low bitrate video compression.
    \item We highlight the importance of semantic conditioning in single-step diffusion and introduce \textit{Contextual Semantic Guidance}, which extracts temporally stable semantics from consecutive frames and refines them through distillation from a pretrained DINOv3 model.
    \item We extend single-step diffusion from image to video reconstruction through \textit{Temporal Consistency Guidance}. Combined with cascade training, TCG enhances temporal alignment and improves overall visual consistency.
    \item Extensive experiments demonstrate that S$^2$VC achieves state-of-the-art compression performance while preserving temporal consistency, showcasing the potential of single-step diffusion for perceptual video compression.
\end{itemize}

%% file: sec_camera/2_related.tex
\begin{figure*}[t]
    \centering
    \includegraphics[width=1.0\linewidth]{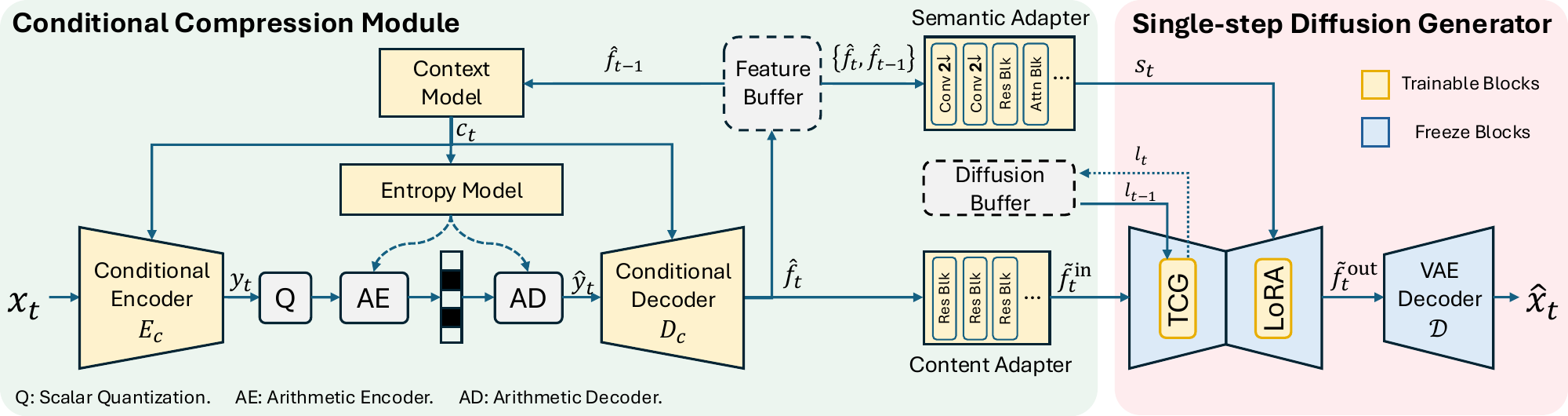}
    \vspace{-7mm}
    \caption{
        Overview of the S$^2$VC framework.
        The feature buffer supports conditional coding, while the diffusion buffer enables feature propagation in TCG blocks for improved temporal consistency.
        LoRA~\cite{hu2022lora} is employed for efficient diffusion fine-tuning.
    }
    \label{fig:3}
    \vspace{-5mm}
\end{figure*}

\section{Related Work}
\label{sec:related}

\subsection{Neural Video Compression}
Neural video compression (NVC) has achieved remarkable progress in recent years, leveraging advances in neural image compression~\cite{balle2018variational, minnen2018joint, wang2023evc} for transform coding and entropy modeling.
The early work DVC~\cite{lu2019dvc} introduced neural residual coding under the low delay reference structure.
Later, DCVC~\cite{li2021deep} shifts the framework into conditional coding, allowing the model to learn more informative temporal context.
Subsequent extensions of DCVC have progressively enhanced condition extraction~\cite{sheng2022temporal}, spatial–temporal entropy modeling~\cite{li2023neural}, and transform coding~\cite{li2024neural}.
The latest DCVC-RT~\cite{jia2025towards} further optimizes the model architecture and runtime efficiency, surpassing the VVC standard~\cite{bross2021overview} while sustaining real-time decoding on consumer-grade hardware.

Despite these advances, existing NVCs are primarily optimized for objective distortion metrics such as MSE or MS-SSIM~\cite{wang2003multiscale}, which often yield oversmoothed results at low bitrate scenarios.  
In contrast, the proposed S$^2$VC targets perceptual quality optimization, achieving more realistic reconstructions at ultra-low bitrates.

\subsection{Perceptual-Oriented Compression}
Perceptual image compression~\cite{mentzer2020high, jia2024generative, xue2025dlf, muckley2023improving} has been widely explored.
HiFiC~\cite{mentzer2020high} integrates perceptual loss with adversarial training, while MS-ILLM~\cite{muckley2023improving} improves generation using a non-binary discriminator.
Recent works leverage pretrained diffusion models~\cite{rombach2021highresolution} to further improve reconstruction.
PerCo~\cite{careil2023towards} fine-tunes diffusion models on vector-quantized image features with generated captions, and DiffEIC~\cite{li2024towards} conditions diffusion models on transform-coded VAE latents with fixed text prompts.
Despite improved quality, their multi-step sampling remains computationally expensive, motivating recent works~\cite{zhang2025stablecodec, xue2025one} to adopt single-step diffusion for image coding.
Another limitation of pretrained text-to-image diffusion models in compression is their use of fixed or caption-based prompts, which provide limited content adaptability or insufficient detail expression.
OneDC~\cite{xue2025one} addresses this by replacing text with content-adaptive hyperprior features, offering more accurate semantic guidance.

Building on progress in perceptual image compression, recent studies have extended these concepts to video coding~\cite{qi2025generative, zheng2026generative}.
Early works~\cite{zhang2021dvc, mentzer2022neural} applied perceptual and adversarial training within neural codecs, while PLVC~\cite{yang2022perceptual} adopted a recurrent autoencoder with a recurrent discriminator to enhance generation realism.
Nevertheless, their limited generation capacity leads to visible artifacts at low bitrates.
Recent works leverage pretrained diffusion models as frame generators to improve quality.
I$^2$VC~\cite{liu20242} and DiffVC~\cite{ma2025diffusion} employ multi-step diffusion with conditional coding to enhance video quality, but they incur substantial computational overhead.
To mitigate this, DiffVC-OSD~\cite{ma2025diffvc} introduces single-step diffusion into video coding.
However, existing diffusion-based codecs still operate at comparatively high bitrates. This motivates the development of single-step diffusion codecs with higher compression ratios, leveraging generative priors in a more efficient manner.

Although both S$^2$VC and OneDC employ enhanced semantic guidance for single-step diffusion, their objectives are different.
OneDC focuses on single-image compression and relies on a vector-quantized hyperprior that preserves only \textit{coarse semantics}.
In contrast, S$^2$VC requires temporally coherent semantics for video, capturing both frame-wise and temporal-aware content. 
Thus, we extract \textit{fine-grained semantics} from continuous buffered features and refine them with a DINOv3~\cite{simeoni2025dinov3} teacher, controlling diffusion together with \textit{Temporal Consistency Guidance}.

\subsection{Diffusion Models}
Image diffusion models have advanced rapidly in recent years.
The Latent Diffusion Model (LDM)~\cite{rombach2021highresolution} performs generation in a compressed latent space, enabling efficient high-resolution synthesis.
Subsequent works~\cite{guo2023animatediff, blattmann2023stable} extend LDMs to video by introducing temporal modules into the U-Net to capture motion dynamics.
To enhance sampling efficiency, recent works distill multi-step diffusion into few-step generators.
Song et al.~\cite{song2023consistency} introduce consistency models for few-step generation, while Yin et al.~\cite{yin2024one, yin2024improved} propose Distribution Matching Distillation (DMD) to align teacher and single-step generated distributions.

Building on those advancements, recent video super-resolution works also exploit single-step diffusion.
UltraVSR~\cite{liu2025ultravsr} utilizes a bidirectional recurrent shift module into the U-Net to handle motion synthesis, while DLoRAL~\cite{sun2025one} employs consistency and detail LoRAs to balance temporal stability and spatial fidelity.
Unlike video SR, which often adopts bidirectional design, our codec follows the causal low-delay structure to enable frame-wise coding.
Moreover, we also consider the joint optimization of the diffusion model and codec, which is an aspect not exists in SR.

%% file: sec_camera/3_method.tex
\section{Methodology}
\label{sec:method}

\subsection{Framework Overview}
We propose S$^2$VC, a Single-Step diffusion–based generative Video Codec that achieves high-quality reconstruction with efficient decoding at low bitrates.
As illustrated in Fig.~\ref{fig:3}, S$^2$VC integrates a conditional compression module with an efficient single-step diffusion generator.
Unlike prior NVCs~\cite{li2024neural, ma2025diffusion, ma2025diffvc} which rely on compressed optical flow for motion-compensated context modeling, our conditional compression module adopts a simplified design~\cite{jia2025towards}.
By removing explicit motion compression, it allows the network to learn inter-frame redundancy implicitly, thereby simplifying both the architecture and the training process.

Starting from the conditional compression module, we first extract temporal context $c_t$ from the previous frame, and encode the current frame into a context-conditioned latent representation: $y_{t} = E_{c}(x_{t}, c_t)$.
The latent representation is subsequently quantized and entropy-coded using probability parameters estimated by the spatial–temporal entropy model~\cite{jia2025towards}, resulting in the compressed latent $\hat{y}_t$.
On the decoder side, the reconstructed feature is obtained from the temporal context: $\hat{f}_{t} = D_{c}(\hat{y}_{t}, c_t)$ and stored in the feature buffer.
The \textit{content adapter} transforms $\hat{f}_t$ into the diffusion latent space, producing $\tilde{f}_t^{\text{in}}$ for single-step generation. 
Meanwhile, the \textit{semantic adapter} used for \textit{Contextual Semantic Guidance} derives the guidance $s_{t}$ from the buffered features $\{\hat{f}_{t}, \hat{f}_{t-1} \}$, serving as a replacement for the text embedding typically used in diffusion models.

Finally, the diffusion generator refines the latent in a single step: $\{ \tilde{f}_{t}^{\text{out}}, l_t \}=\epsilon_{\theta}(\tilde{f}_{t}^{\text{in}}, s_{t}, l_{t-1})$, where $l_t$ denotes the intermediate features propagated through the \textit{Temporal Consistency Guidance} blocks within the U-Net.
This refined latent is then decoded by a pretrained VAE decoder to yield the high-quality reconstruction: $\hat{x}_{t} = \mathcal{D}(\tilde{f}_{t}^{\text{out}})$.
With the proposed semantic–temporal guidance, we adapt the image diffusion generator into a powerful video generator.
The two forms of guidance are detailed in the following sections.

\subsection{Contextual Semantic Guidance}
\noindent \textbf{Preliminary.}
Pretrained diffusion models are typically designed for text-to-image generation, using text embeddings as semantic guidance.
In compression tasks, prior works often rely on fixed prompts or generated captions, which either lack adaptability or fail to convey detailed spatial semantics.
To overcome this, OneDC~\cite{xue2025one} replaces text embeddings with semantic representation $s$ extracted from hyperprior features~\cite{balle2018variational}, enabling spatially localized and content-aware conditioning.
In each cross-attention layer of the U-Net, the input feature $f_{\text{in}}'$ serves as the query, while the semantic feature $s$ supplies the keys and values:

\vspace{-3mm}
\begin{equation}
\label{eq:0}
\begin{aligned}
f_\text{out}' &= \mathrm{Softmax}\!\left(\frac{QK^\top}{\sqrt{d_{k}}}\right)V, \quad\text{where}\\
Q &= W_{Q}f_\text{in}', ~ K = W_{K}s, ~ V = W_{V}s
\end{aligned}
\end{equation}
\vspace{-3mm}

\noindent where $d_k$ is the dimensionality of the key vectors and $f_\text{out}'$ denotes the output feature.
For stronger semantic representation, OneDC further transfers tokenizer priors into the hyperprior branch through distillation.

However, in conditional video coding, the hyperprior mainly captures the distribution of inter-frame differences rather than actual image content, making it unsuitable for guidance extraction.
In addition, the temporal dimension in video requires semantic guidance that represents temporal dynamics rather than merely frame-wise information, while remains stable across frames to ensure consistent reconstruction.
These factors indicate that the coarse semantics used in OneDC are inadequate for video, motivating S$^2$VC to adopt a more advanced guidance design.

\vspace{1mm}
\noindent \textbf{Semantic Guidance from Context.}
In conditional video coding, the decoded feature buffer inherently contains rich per-frame representations.
We therefore introduce \textit{Contextual Semantic Guidance (CSG)} to extract such guidance $s_{t}$ from this buffer.
Concretely, the \emph{Semantic Adapter} takes the buffered pair $\{\hat{f}_{t}, \hat{f}_{t-1}\}$ as input and applies a series of strided convolutions, residual blocks, and attention layers for spatial–temporal aggregation (see Fig.~\ref{fig:3}), yielding accurate and temporally stable semantic features.
Through end-to-end training, the resulting guidance $s_{t}$ is aligned with the diffusion conditioning space, providing fine-grained, content-adaptive guidance as formulated in Eq.~\ref{eq:0}.

\begin{figure}[t]
    \centering
    \includegraphics[width=1.0\linewidth]{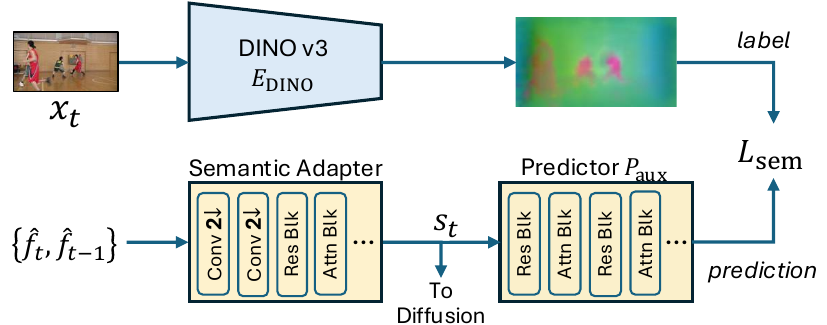}
    \vspace{-6mm}
    \caption{
        Semantic Distillation in S$^2$VC.
        The pretrained DINOv3 serves as a teacher, providing temporally stable and semantically rich features that are distilled into the semantic adapter.
    }
    \label{fig:4}
    \vspace{-4mm}
\end{figure}

Notably, the \emph{Semantic Adapter} complements the \textit{Content Adapter}, as they serve different purposes: the former focuses on a lower-resolution, temporally aggregated abstraction (leveraging both $\hat{f}_{t-1}$ and $\hat{f}_{t}$) to provide high-level information, whereas the latter maps the current feature $\hat{f}_{t}$ into the diffusion latent for pixel-level reconstruction.
Decoupling these roles yields stable semantics for guidance while retaining high-fidelity detail synthesis.

\vspace{1mm}
\noindent \textbf{Semantic Distillation.}
To enable the CSG module to extract temporally stable and semantically rich features, we introduce a semantic distillation scheme.
The recent DINOv3 model~\cite{simeoni2025dinov3} produces representations that are both temporally consistent and semantically expressive, and has demonstrated strong effectiveness in diffusion training~\cite{leng2025repa}.
Inspired by this, we adopt DINOv3 as a teacher network and transfer its semantic capability to our \textit{Semantic Adapter}.

As shown in Fig.~\ref{fig:4}, an auxiliary predictor $P_\text{aux}$ is introduced to map the learned semantic representation $s_{t}$ into the DINOv3 feature space.
The predictor is optimized using an $L_1$ loss, and gradients are jointly propagated to both the CSG module and the conditional codec:
\begin{equation}
    \label{eq:1}
    L_\text{sem}=||E_{\text{DINO}}(x_t) - P_{\text{aux}}(s_{t})||_{1}
\end{equation}

Both $P_\text{aux}$ and $E_{\text{DINO}}$ are used only during training, incurring no additional overhead at inference.
As shown in Sec.~\ref{sec:ablation}, ablation results confirm that this distillation significantly enhances reconstruction quality, further validating the importance of semantic guidance and distillation.

\begin{figure}[t]
    \centering
    \includegraphics[width=1.0\linewidth]{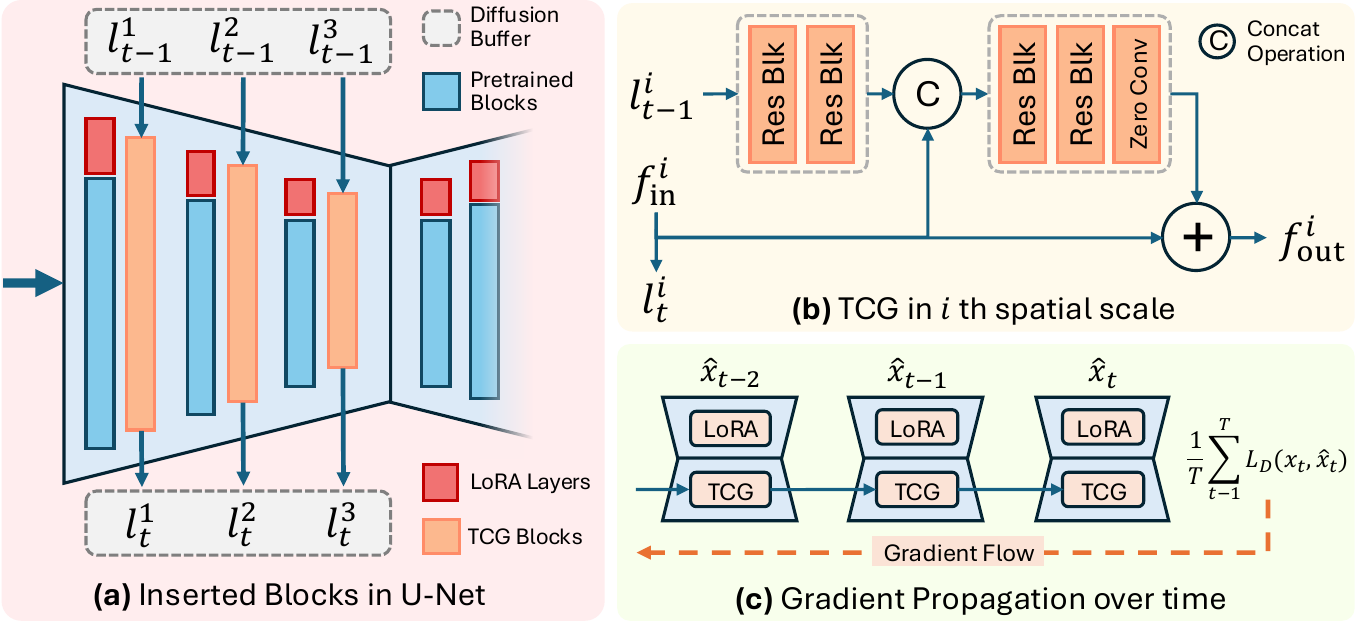}
    \vspace{-6mm}
    \caption{
        Temporal Consistency Guidance (TCG) design and gradient flow.
        $l^{i}_{t}$ denotes the intermediate feature at the $i$-th scale of frame $t$.
        $f^{i}_{\text{in}}$ / $f^{i}_{\text{out}}$ are the input and output of a TCG block.
        $L_{D}$ is the distortion loss, and $T$ is the total frame count.
    }
    \label{fig:5}
    \vspace{-4mm}
\end{figure}

\subsection{Temporal Consistency Guidance}
In S$^2$VC, we employ an image diffusion model for frame-wise causal coding, following recent diffusion-based video codecs~\cite{ma2025diffusion, ma2025diffvc}.
However, achieving high perceptual quality in video compression requires not only superior frame-level reconstruction but also strong temporal consistency across frames.
Therefore, a key challenge lies in effectively extending the image diffusion model to the video domain while preserving such temporal coherence.

To this end, we propose the \textit{Temporal Consistency Guidance (TCG)} design, a set of plug-in blocks integrated into the diffusion U-Net.
As shown in Fig.~\ref{fig:5}(a, b), multiple TCGs are inserted into the encoder part of the U-Net at different scales.
Each TCG retrieves the intermediate features of the previous frame from the diffusion buffer, and fuses them with the corresponding features of the current frame, enabling temporal propagation.
The input feature is then written back to the buffer and reused in subsequent frame coding.
This design offers two advantages: (1) TCGs enable adaptive reuse of previously synthesized textures while integrating new content in the current frame through end-to-end training. (2) The plug-in design with zero-conv initialization ensures that TCGs introduce temporal modeling capability without harming the pretrained generative prior.

While TCG propagates features across frames, another challenge lies in optimizing it to sufficiently exploit temporal correlations for consistent video generation.
Inspired by the cascade training used in the DCVC series~\cite{li2023neural, li2024neural, jia2025towards}, we extend this mechanism to our single-step diffusion framework, as illustrated in Fig.~\ref{fig:5}(c).
Concretely, the gradients of the current frame’s intermediate features are propagated backward to previous frames, forming a temporal optimization chain.
The resulting chain effectively captures inter-frame correlations within the latent representations, producing more accurate and temporally consistent reconstructions, as confirmed by our experiments in Sec.~\ref{sec:ablation}.

\begin{figure*}[t]
    \centering
    \includegraphics[width=1.0\linewidth]{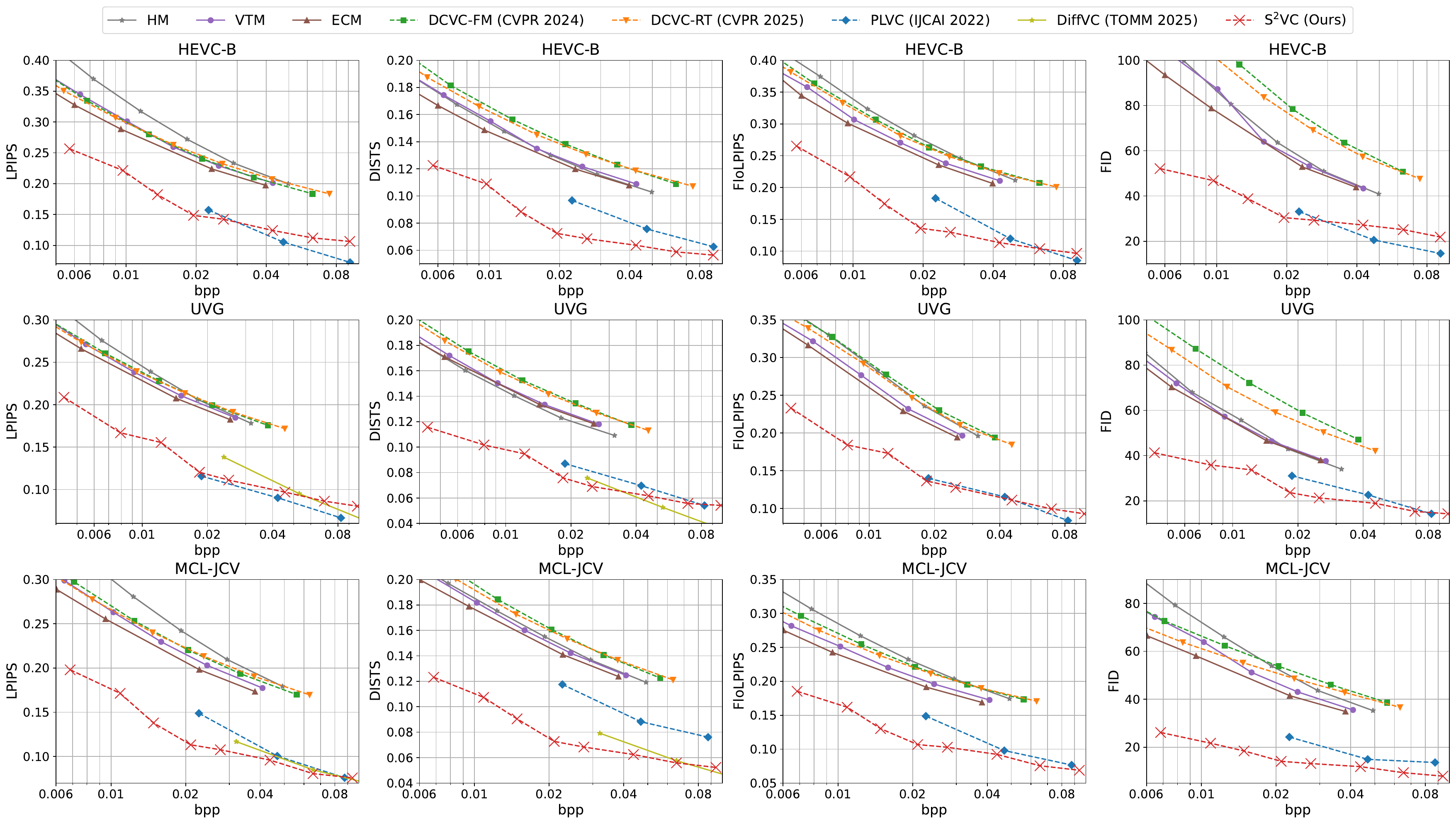}
    \vspace{-7mm}
    \caption{
        The rate-perception curves of the S$^2$VC and other video compression methods.
    }
    \label{fig:exp1}
    \vspace{-5mm}
\end{figure*}

%% file: sec_camera/4_experiment.tex
\section{Experiment}
\label{sec:experiment}

\subsection{Implementation Details}

\noindent \textbf{Pretrained Models.}
We employ the OneDC~\cite{xue2025one} image codec to compress the I-frame.
The parameters of the single-step diffusion in S$^2$VC are initialized from the DMD2 SD1.5 model~\cite{yin2024improved}.
For semantic supervision, we adopt the distilled ConvNeXt variant of DINOv3~\cite{simeoni2025dinov3} as the teacher network, which supports variable input resolutions.

\vspace{1mm}
\noindent \textbf{Training settings.} 
To train S$^2$VC in an end-to-end manner, we adopt a perceptually oriented RD loss defined as:
\begin{equation}
    L = \frac{1}{T}\sum_{t=1}^{T}(\lambda R + L_{\text{D}} + \alpha L_\text{sem} + \beta L_\text{motion})
\end{equation}
\noindent where $L_{\text{D}}$ denotes the distortion loss, $R$ represents the bitrate estimated by the spatial–temporal entropy model~\cite{jia2025towards}, and $T$ is the total number of frames in the sequence.
The parameter $\lambda$ controls the rate–distortion trade-off.
$L_{\text{sem}}$ is the semantic distillation loss (Eq.~\ref{eq:1}), and $L_{\text{motion}}$ is the optical-flow–based motion loss that enhances temporal consistency.

Specifically, the $L_{\text{D}}$ consists of a pixel-wise $L_1$ term and an LPIPS term. The $L_{\text{motion}}$ minimizes the $L_1$ distance between the optical flows computed from consecutive original and reconstructed frames using a pretrained RAFT model $O$~\cite{teed2020raft}, thereby enhancing temporal consistency:
\begin{align}
    L_{\text{D}} &= ||x_t-\hat{x}_t||_{1} + L_\text{LPIPS}(x_t, \hat{x}_t) \\
    L_\text{motion} &= ||O(x_{t-1}, x_{t}) - O(\hat{x}_{t-1}, \hat{x}_{t})||_{1}
\end{align}
We optimize the model with the AdamW~\cite{loshchilov2017decoupled} optimizer, applying a multi-stage schedule for learning rate and sequence length.
The single-step diffusion model is adapted via LoRA layers~\cite{hu2022lora}, allowing fast convergence for the compression while preserving rich generative priors.
Additional details are provided in the supplementary material.

\vspace{1mm}
\noindent \textbf{Evaluation settings.}
Following~\cite{li2023neural, li2024neural, ma2025diffusion}, we evaluate the first 96 frames of each sequence under the low-delay setting, with one I-frame followed by P-frames.
For perceptual quality assessment, we adopt frame-wise LPIPS~\cite{johnson2016perceptual} and DISTS~\cite{ding2020image}.
To evaluate temporal consistency, we use FloLPIPS~\cite{danier2022flolpips}, a motion-aware perceptual metric that measures inter-frame coherence.
For evaluating generation realism, we compute FID~\cite{heusel2017gans} on overlapping $256\times256$ patches extracted from video frames, following prior practice~\cite{mentzer2020high}.

\vspace{1mm}
\noindent \textbf{Datasets.}
For model training, we use the OpenVid-HD dataset~\cite{nan2024openvid}, which provides video clips with high aesthetics, clarity, and resolution suitable for generative model training.
For model evaluation, we adopt the HEVC-B~\cite{sullivan2012overview}, UVG~\cite{mercat2020uvg}, and MCL-JCV~\cite{wang2016mcl} datasets, all containing 1920$\times$1080 videos that closely reflect practical scenarios.

\vspace{1mm}
\noindent \textbf{Compared Methods.}
We conduct a comprehensive comparison between S$^2$VC and several publicly available video codecs, including:
(1) Traditional reference software: HM~\cite{sullivan2012overview}, VTM~\cite{bross2021overview}, and ECM~\cite{jvet2025ecm};
(2) Distortion-oriented neural codecs: DCVC-FM~\cite{li2024neural} and DCVC-RT~\cite{jia2025towards};
and (3) Perception-oriented neural codec: PLVC~\cite{yang2022perceptual}.
We also report the comparison of the publicly unavailable diffusion-based codec DiffVC~\cite{ma2025diffusion}, by extracting their reported data.

\begin{figure*}[t]
    \centering
    \includegraphics[width=1.0\linewidth]{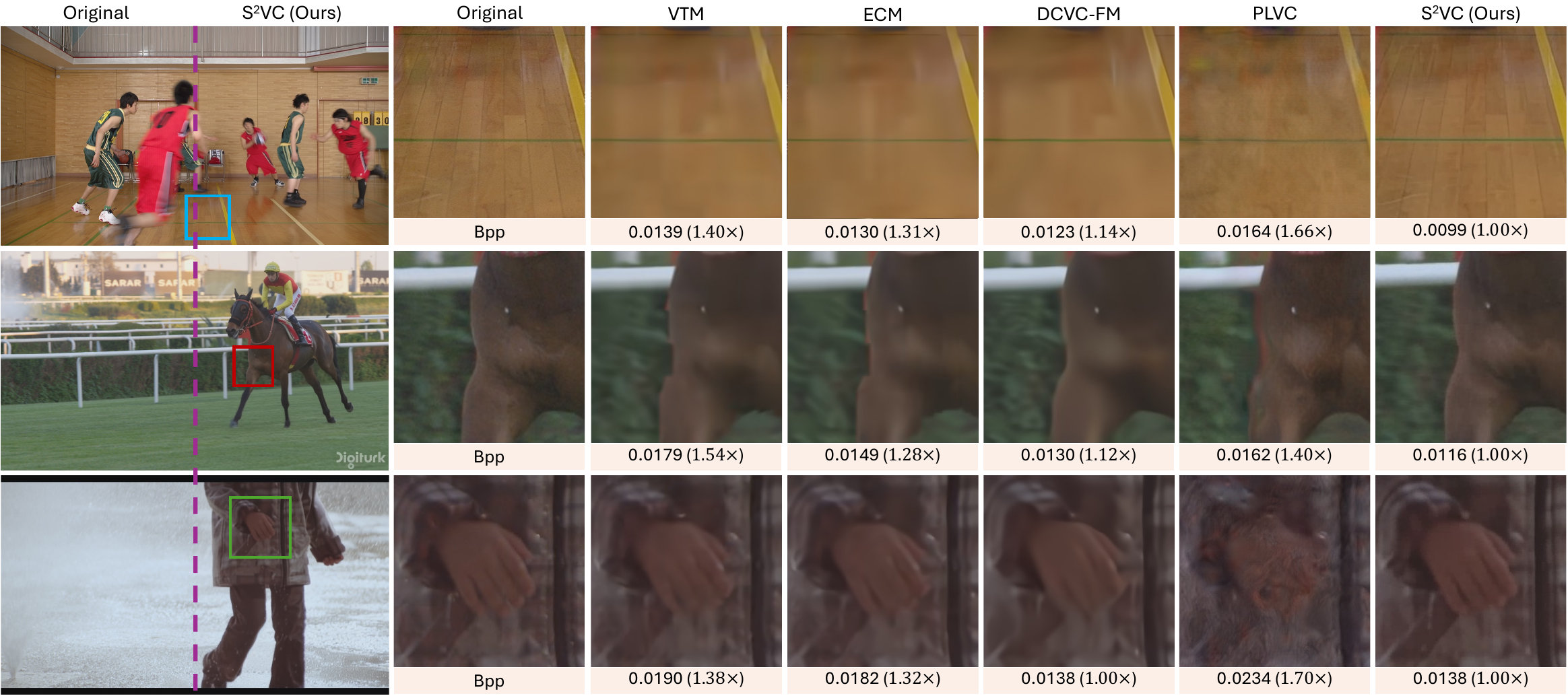}
    \vspace{-7mm}
    \caption{
        Visual comparisons.
        \textit{Top:} complex motion; \textit{Mid:} panning with moving objects; \textit{Bottom:} small motion.
        Zoom in for better view.
    }
    \label{fig:exp2}
    \vspace{-4mm}
\end{figure*}

\subsection{Experimental Results}
\noindent \textbf{Quantitative Evaluation.}
Fig.~\ref{fig:exp1} illustrates the comparison between S$^2$VC and existing methods across multiple distortion metrics.
S$^2$VC achieves a clear performance advantage in both fidelity and realism-oriented metrics across all datasets while maintaining the lowest bitrate.
Compared with traditional codecs and distortion-oriented neural codecs (e.g., ECM and DCVC), S$^2$VC exhibits substantially lower DISTS and LPIPS, demonstrating its superior perceptual quality under extreme compression.
When compared with the latest publicly available perceptual codec PLVC, S$^2$VC achieves an average bitrate saving of \textbf{51.62\%} under the DISTS metric, with detailed savings of \textbf{57.69\%}, \textbf{32.69\%}, and \textbf{64.49\%} on the HEVC-B, UVG, and MCL-JCV datasets, respectively (calculated using BD-Rate~\cite{bjontegaard2001calculation}).
Meanwhile, it attains comparable or superior results under the LPIPS metric, demonstrating strong frame-level perceptual fidelity.
For the motion-aware FloLPIPS metric, S$^2$VC outperforms compared codecs on the HEVC-B and MCL-JCV datasets, which feature extensive motion scenarios, demonstrating strong robustness in maintaining temporal consistency.
Furthermore, S$^2$VC attains the best FID across all datasets, confirming that the diffusion prior effectively enhances the realism and naturalness of the reconstructed videos.
Additional results (PSNR and MS-SSIM) are provided in the supplementary material.

\begin{figure}[t]
    \centering
    \includegraphics[width=1.0\linewidth]{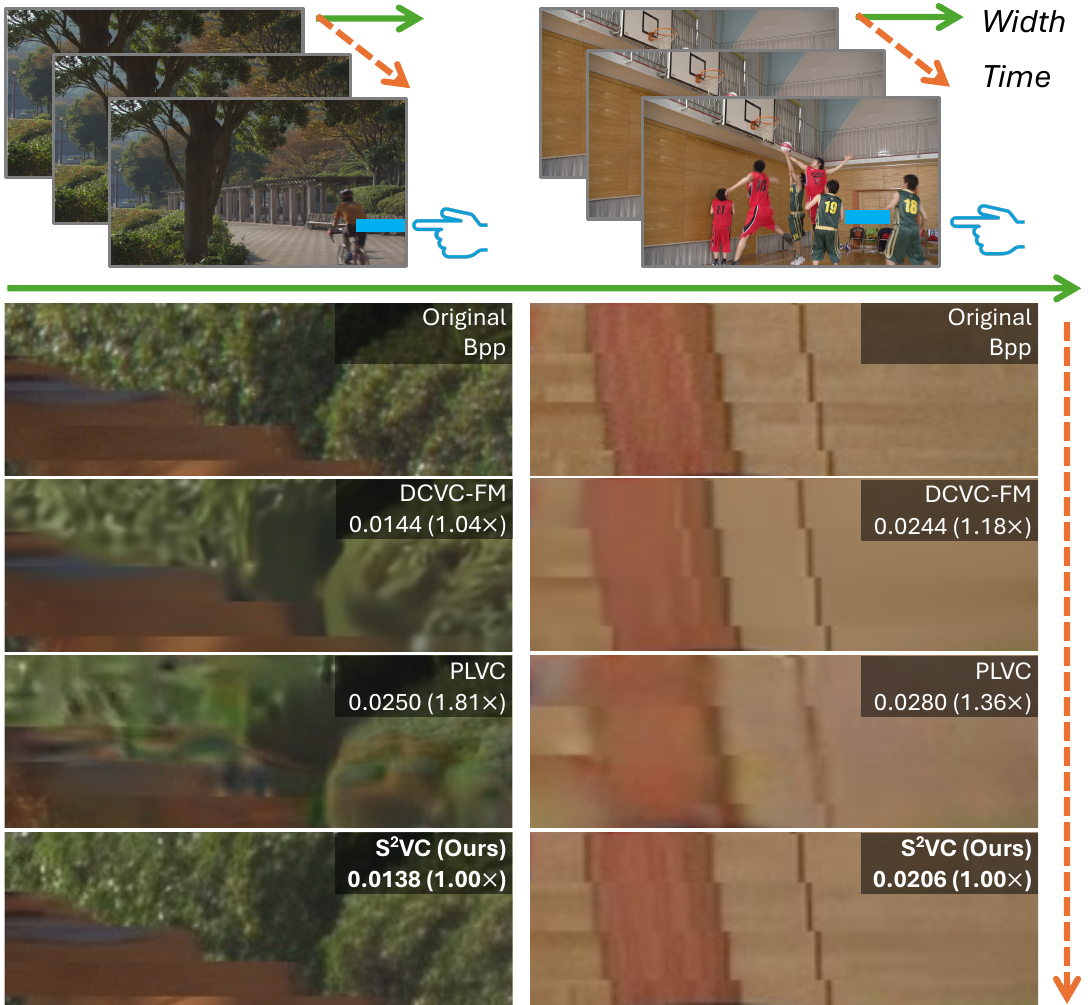}
    \vspace{-7mm}
    \caption{
        Comparison of temporal consistency by stacking the \textcolor{blue}{blue line} across frames.
        The misaligned lines are caused by camera motion.
        Results closer to the original imply better temporal consistency and visual quality.
        Zoom in for better view.
    }
    \label{fig:exp3}
    \vspace{-5mm}
\end{figure}

\noindent \textbf{Qualitative Evaluation.}
Fig.~\ref{fig:exp2} presents visual comparisons across diverse scenes, including regions with complex motion (top), panning backgrounds with moving objects (middle), and small-scale object motions (bottom).
Traditional codecs such as VTM and ECM produce blurry reconstructions in all examples, with evident blocking and ringing artifacts in motion edges (particularly in the middle example).
The neural codec DCVC-FM, though free from blocking artifacts, still yields over-smoothed details and degraded visual quality.
While PLVC improves detail sharpness, it introduces noticeable variegated and mottled artifacts, particularly in motion regions (middle and bottom).
In contrast, the proposed S$^2$VC faithfully reconstructs fine details with high perceptual fidelity, even under complex motion.
Notably, it recovers floor textures despite fast-moving subjects (top), preserves accurate edges and details on the horse (middle), and clearly delineates all fingers (bottom).

\begin{table*}[t]
\captionof{table}{\small Ablation studies calculated by BD-Rate ($\%$) $\downarrow$. \textbf{Ours} is the anchor (0.00\%).}
\label{tab:1}
\vspace{-2mm}
\centering
\resizebox{0.85\textwidth}{!}{
\begin{tabular}{@{}lcccc|cccc@{}}
\toprule
\multicolumn{1}{c}{Dataset} & \multicolumn{4}{c}{HEVC-B} & \multicolumn{4}{c}{UVG} \\[0.0em] \cmidrule(l){2-9} 
\multicolumn{1}{c}{\small Metrics for BD-Rate Calculation} & LPIPS & DISTS & FloLPIPS & FID & LPIPS & DISTS & FloLPIPS & FID \\[0.05em] \midrule
\textcolor{gray}{\emph{Contextual Semantic Guidance (CSG)}} \\ 
w/o CSG & 27.46 & 22.08 & 25.65 & 28.05 & 43.75 & 32.16 & 39.32 & 18.36 \\
w/ CSG only & 13.30 & 14.06 & 14.10 & 7.28 & 34.00 & 29.71 & 28.38 & 13.97 \\
w/ CSG + distill $\to$ \textbf{Ours} & 0.00 & 0.00 & 0.00 & 0.00 & 0.00 & 0.00 & 0.00 & 0.00 \\ \midrule
\textcolor{gray}{\emph{Temporal Consistency Guidance (TCG)}} \\ 
w/o TCG & 20.41 & 23.67 & 28.13 & 9.54 & 38.75 & 26.70 & 46.88 & 4.87 \\
w/ TCG only & 11.74 & 11.25 & 14.66 & 6.32 & 33.92 & 21.92 & 28.46 & 1.96 \\
w/ TCG + cascade $\to$ \textbf{Ours} & 0.00 & 0.00 & 0.00 & 0.00 & 0.00 & 0.00 & 0.00 & 0.00 \\ \bottomrule

\end{tabular}
}
\vspace{-2mm}
\end{table*}

\begin{figure}
    \centering
    \includegraphics[width=1.0\linewidth]{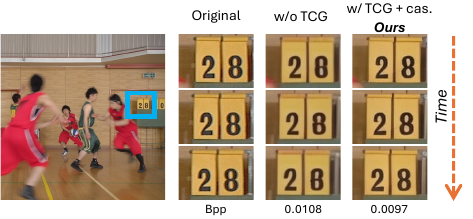}
    \vspace{-6mm}
    \caption{
        Ablation study of Temporal Consistency Guidance.
        Removing TCG causes distortion in edges of the digit "8".
    }
    \label{fig:exp5}
    \vspace{-4mm}
\end{figure}

Fig.~\ref{fig:exp3} further compares temporal consistency.
DCVC-FM exhibits severe blurring in regions where objects move across frames, while PLVC introduces noticeable jitter artifacts on moving objects, leading to an unstable visual appearance.
In contrast, S$^2$VC preserves per-frame fidelity and ensures smooth motion continuity without introducing artifacts, while effectively restoring complex patterns in regions traversed by moving objects.
This further validates high perceptual quality and superior temporal consistency.

\subsection{Ablation Studies}
\label{sec:ablation}
We conduct ablation studies to investigate the effectiveness of the Contextual Semantic Guidance (CSG) and the proposed Temporal Consistency Guidance (TCG) designs. Models are evaluated on the HEVC-B~\cite{sullivan2012overview} and UVG~\cite{mercat2020uvg} datasets with BD-Rate~\cite{bjontegaard2001calculation} on various metrics.

\vspace{1mm}
\noindent \textbf{Contextual Semantic Guidance.}
We first validate the importance of CSG in the single-step diffusion model and then assess the effectiveness of semantic distillation using the DINOv3 teacher.
As shown in Table~\ref{tab:1}, removing the semantic guidance input for the single-step diffusion model (“w/o CSG”) significantly degrades reconstruction quality across all metrics, highlighting the crucial role of semantic guidance in enhancing generation performance.
Introducing contextual semantic guidance (“w/ CSG only”) notably improves reconstruction results, demonstrating that features extracted from contextual frames provide meaningful guidance even without external supervision.
Further applying semantic distillation from the DINOv3 teacher (“w/ CSG + distill”) yields additional improvements, confirming that transferring prior knowledge from a pretrained vision model enhances the semantic accuracy of the guidance module.

\vspace{1mm}
\noindent \textbf{Temporal Consistency Guidance.}
We remove all TCG blocks from the diffusion U-Net (“w/o TCG”) to evaluate the impact of temporal guidance.
As shown in Table~\ref{tab:1}, this variant shows a clear performance drop, particularly in FloLPIPS, indicating degraded temporal consistency and highlighting the importance of the TCG design.
This degradation is also evident in Fig.~\ref{fig:exp5}, where the edges of the digit exhibit noticeable jitter and distortion after removing the TCG blocks.
To further analyze the role of cascade training, we retain TCG blocks but disable gradient propagation across frames (“w/ TCG only”), which slightly improves quality.
When cascade training is applied (“w/ TCG + cascade”), additional gains are achieved. 
This result confirms that TCG is crucial for temporal consistency and cascade training effectively maximizes its benefit.

\section{Conclusion}
We propose S$^2$VC, a single-step diffusion–based video codec that achieves high perceptual quality and temporal consistency at ultra-low bitrates.
By integrating a conditional coding framework with a diffusion generator, S$^2$VC effectively leverages generative priors to reconstruct high-fidelity details while improving compression ratio.
The proposed \textit{Contextual Semantic Guidance} provides content-adaptive and temporally stable semantics distilled from DINOv3, enabling the diffusion generator to reconstruct faithful details without external captioning or prompts.
Meanwhile, the \textit{Temporal Consistency Guidance} aligns features across frames through multi-scale fusion and cascade optimization, effectively suppressing flicker and ensuring coherent motion.
Together, they allow the diffusion generator to faithfully restore fine details without jitter or blur, achieving SOTA perceptual quality and temporal consistency under extreme bitrate constraints.

\vspace{1mm}
\noindent \textbf{Limitation.} 
While S$^2$VC achieves high-quality and temporally stable reconstruction with single-step diffusion sampling, its current operating bitrate range remains limited. Future work will focus on architectural refinement to support a broader range of compression ratios.

%% file: sec_camera/5_suppl.tex

\section{Experiment}
This section first outlines the detailed experimental settings and then presents additional quantitative / qualitative results for a more comprehensive evaluation.

\subsection{Compared Methods}
Detailed configurations for each method are listed below.

\vspace{1mm}
\noindent \textbf{Traditional Video Codecs.}
We evaluate VTM-17.0~\cite{bross2021overview} and ECM-5.0~\cite{jvet2025ecm} in RGB color space, following the setting used in DCVC series~\cite{li2023neural}, where RGB input is converted to 10-bit YUV444 for internal codec processing.
As noted in~\cite{li2023neural}, this configuration improves compression performance.
The following configuration files are used for each codec:

\begin{itemize}
    \item VTM: \textit{encoder\_lowdelay\_vtm.cfg}
    \item ECM: \textit{encoder\_lowdelay\_ecm.cfg}
\end{itemize}

\noindent The parameters used for each video are as follows:

\begin{itemize}
    \item \texttt{--}c \{config file name\} \\ 
    \texttt{--}InputFile=\{input video name\} \\
    \texttt{--}InputBitDepth=10 \\
    \texttt{--}OutputBitDepth=10 \\
    \texttt{--}OutputBitDepthC=10 \\
    \texttt{--}InputChromaFormat=444 \\
    \texttt{--}FrameRate=\{frame rate\} \\
    \texttt{--}DecodingRefreshType=2 \\
    \texttt{--}FramesToBeEncoded=\{frame number\}  \\
    \texttt{--}SourceWidth=\{width\}  \\
    \texttt{--}SourceHeight=\{height\}  \\
    \texttt{--}IntraPeriod=\{intra period\}  \\
    \texttt{--}QP=\{qp\}  \\
    \texttt{--}Level=6.2  \\
    \texttt{--}BitstreamFile=\{bitstream file name\}
\end{itemize}

\vspace{1mm}
Notably, all coding tools and reference structure of traditional codecs use their best settings to achieve optimal compression performance.

\vspace{1mm}
\noindent \textbf{MSE-optimized Neural Codecs.}
We use the officially released models of DCVC-FM~\cite{li2024neural} and DCVC-RT~\cite{jia2025towards}.

\vspace{1mm}
\noindent \textbf{Perceptual-optimized Neural Codecs.}
For PLVC~\cite{yang2022perceptual}, we use the officially released model.
For DiffVC~\cite{ma2025diffusion}, whose implementation is not publicly available, we adopt the results reported in their paper, as their evaluation protocol aligns with ours.

\begin{figure*}[t]
    \centering
    \includegraphics[width=1.0\linewidth]{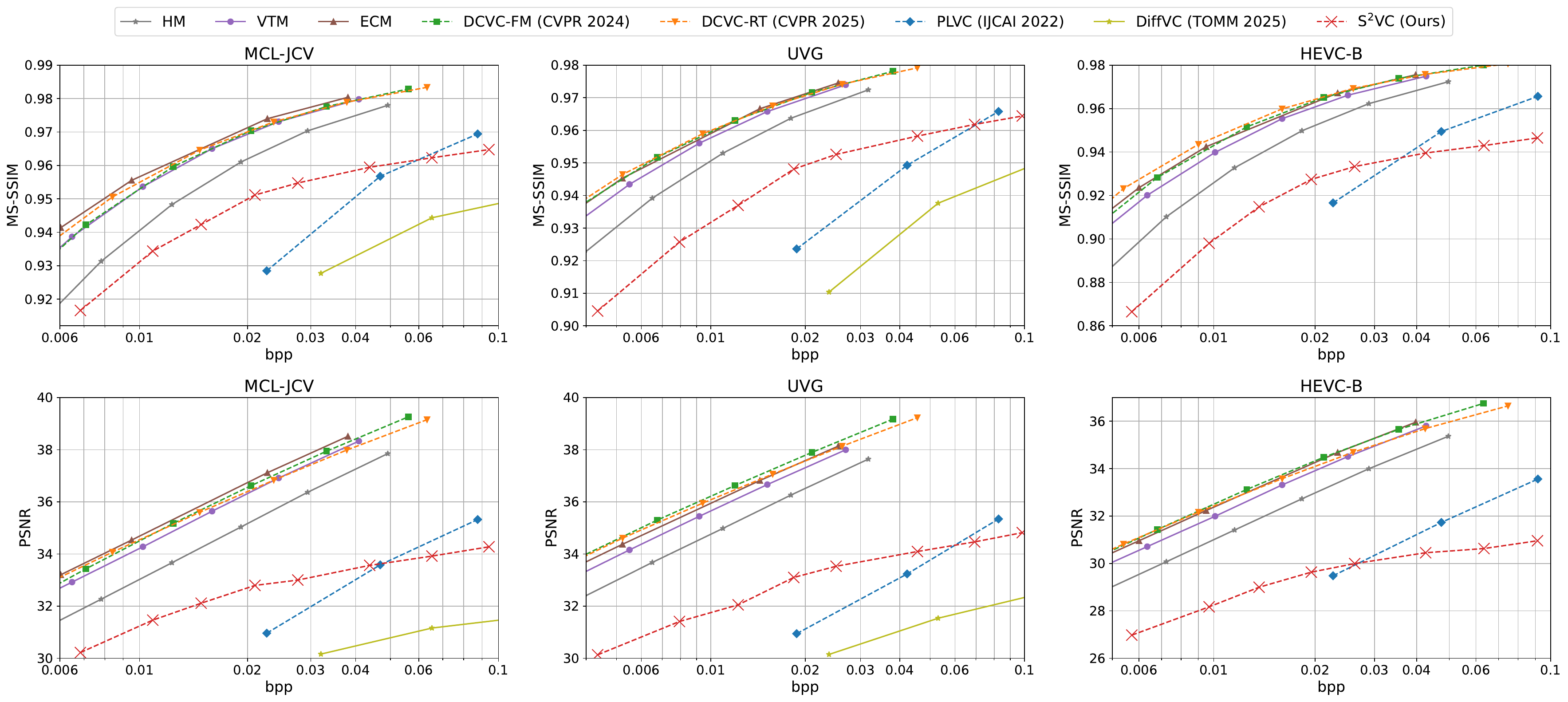}
    \vspace{-4mm}
    \caption{Rate-distortion curves in terms of PSNR and MS-SSIM~\cite{wang2003multiscale}. Datasets: MCL-JCV~\cite{wang2016mcl}, UVG~\cite{mercat2020uvg} and HEVC-B~\cite{sullivan2012overview}.}
    \label{fig:supp_exp1}
    \vspace{-1mm}
\end{figure*}

\begin{figure}[t]
    \centering
    \includegraphics[width=1.0\linewidth]{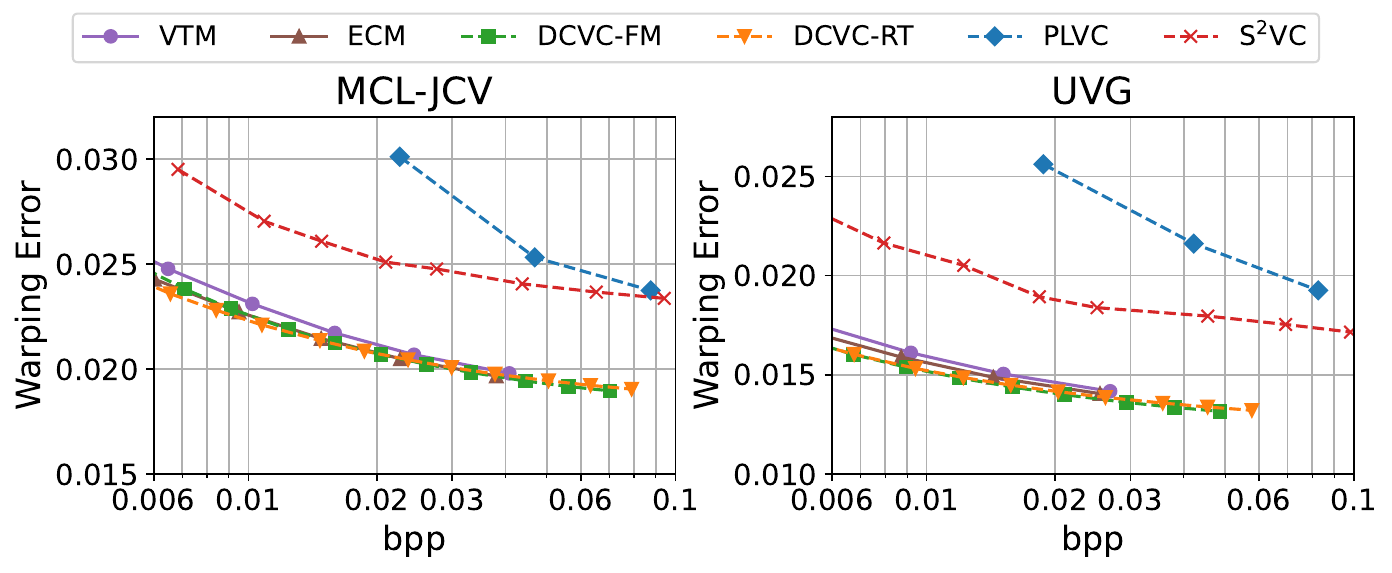}
    \vspace{-6mm}
    \caption{Rate-distortion curves in terms of warping error $E_\text{warp}$.}
    \vspace{-4mm}
    \label{fig:supp_exp_warp}
\end{figure}

\subsection{Additional Quantitative Evaluation}
For completeness, we further evaluate S$^2$VC using PSNR and MS-SSIM~\cite{wang2003multiscale}, as shown in Fig.~\ref{fig:supp_exp1}.
We also report the warping error $E_\text{warp} = \frac{1}{N-1}\sum^{N-1}_{i=1}||x_{i+1}-F_\text{wp}(\hat{x}_{i})||_{1}$ in Fig.~\ref{fig:supp_exp_warp}, where the optical flow used in $F_\text{wp}$ is estimated from GT frames.
At low bitrates, optimizing for PSNR tends to suppress high-frequency details, resulting in overly smooth reconstructions~\cite{mentzer2020high}, as confirmed by our visual examples.
As a result, higher objective metrics at these low bitrates do not necessarily reflect accurate perceptual quality.

Although S$^2$VC produces lower PSNR than objective-only codecs, both perceptual metrics and qualitative results consistently demonstrate its superior visual fidelity.
Furthermore, compared to perceptually optimized codecs like PLVC and DiffVC, S$^2$VC achieves significantly higher objective scores while maintaining better perceptual quality.
These results highlight that S$^2$VC delivers superior fidelity over prior perceptual codecs, without sacrificing visual realism, emphasizing the effectiveness of our approach.

\begin{figure*}[t]
    \centering
    \includegraphics[width=1.0\linewidth]{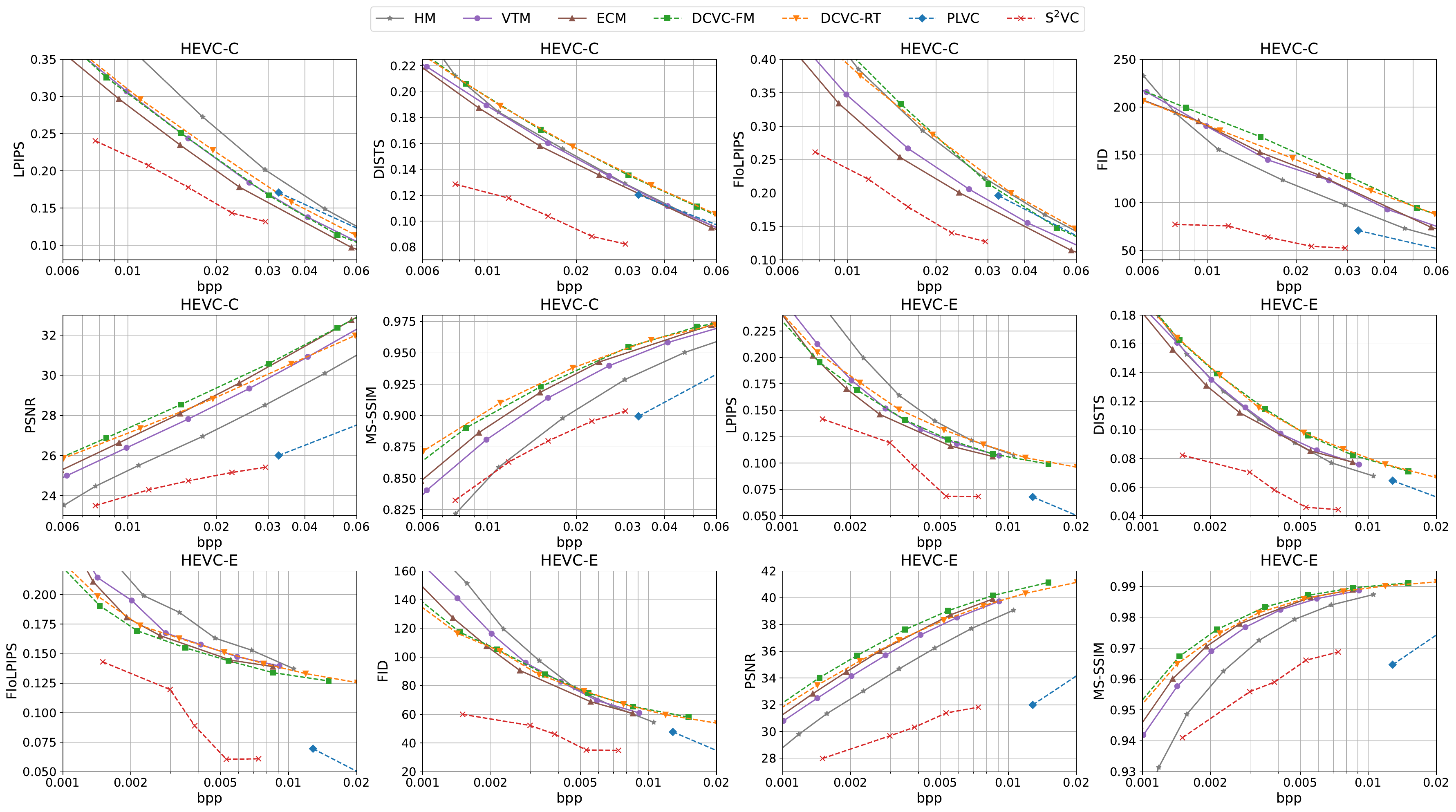}
    \vspace{-6mm}
    \caption{Rate-distortion curves on low-resolution datasets: HEVC-C and HEVC-E~\cite{sullivan2012overview}.}
    \label{fig:supp_exp2}
    \vspace{-3mm}
\end{figure*}

We also report results on the lower-resolution datasets HEVC-C (832$\times$480) and HEVC-E (1280$\times$720), as shown in Fig.~\ref{fig:supp_exp2}.
On these datasets, S$^2$VC continues to outperform other methods in terms of perceptual quality, as measured by LPIPS, DISTS, FloLPIPS, and FID, demonstrating its generalization capability to 480p and 720p videos.

\subsection{Additional Qualitative Examples}
\begin{figure*}[t]
    \centering
    \includegraphics[width=1.0\linewidth]{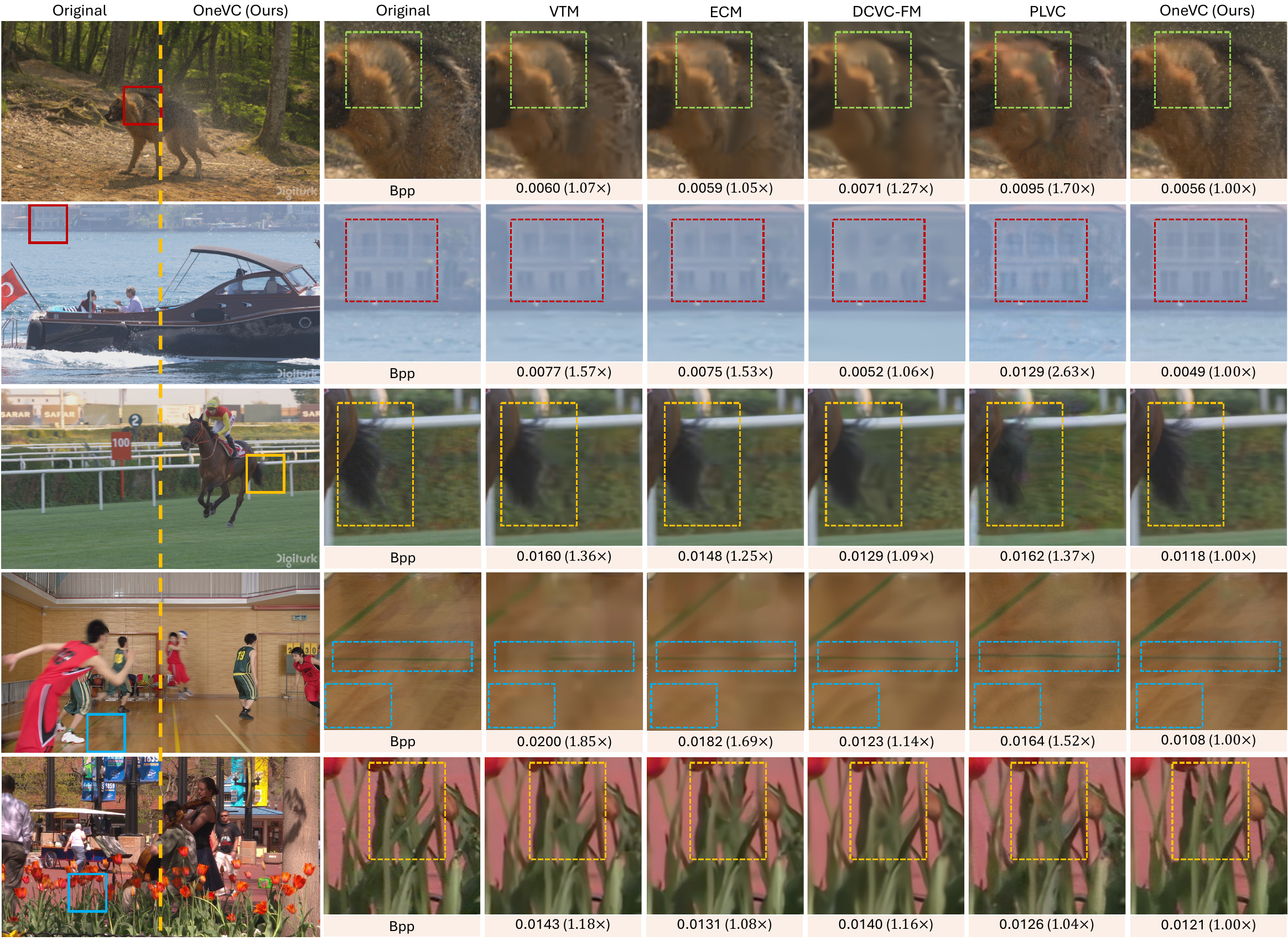}
    \vspace{-5mm}
    \caption{
        Additional visual examples. 
        The proposed S$^2$VC delivers the most realistic and faithful reconstruction at the lowest bitrate. In contrast, traditional codecs (VTM/ECM) and the neural codec DCVC-FM produce blurry reconstructions at low bitrates, while the perceptual codec PLVC generates distorted detail.
    }
    \vspace{-2mm}
    \label{fig:supp_visual_1}
\end{figure*}

We present additional qualitative comparisons on Fig.~\ref{fig:supp_visual_1}.
S$^2$VC consistently outperforms prior video codecs, delivering superior visual quality across diverse content, yet with the lowest bitrate cost.

\begin{figure*}[t]
    \centering
    \includegraphics[width=1.0\linewidth]{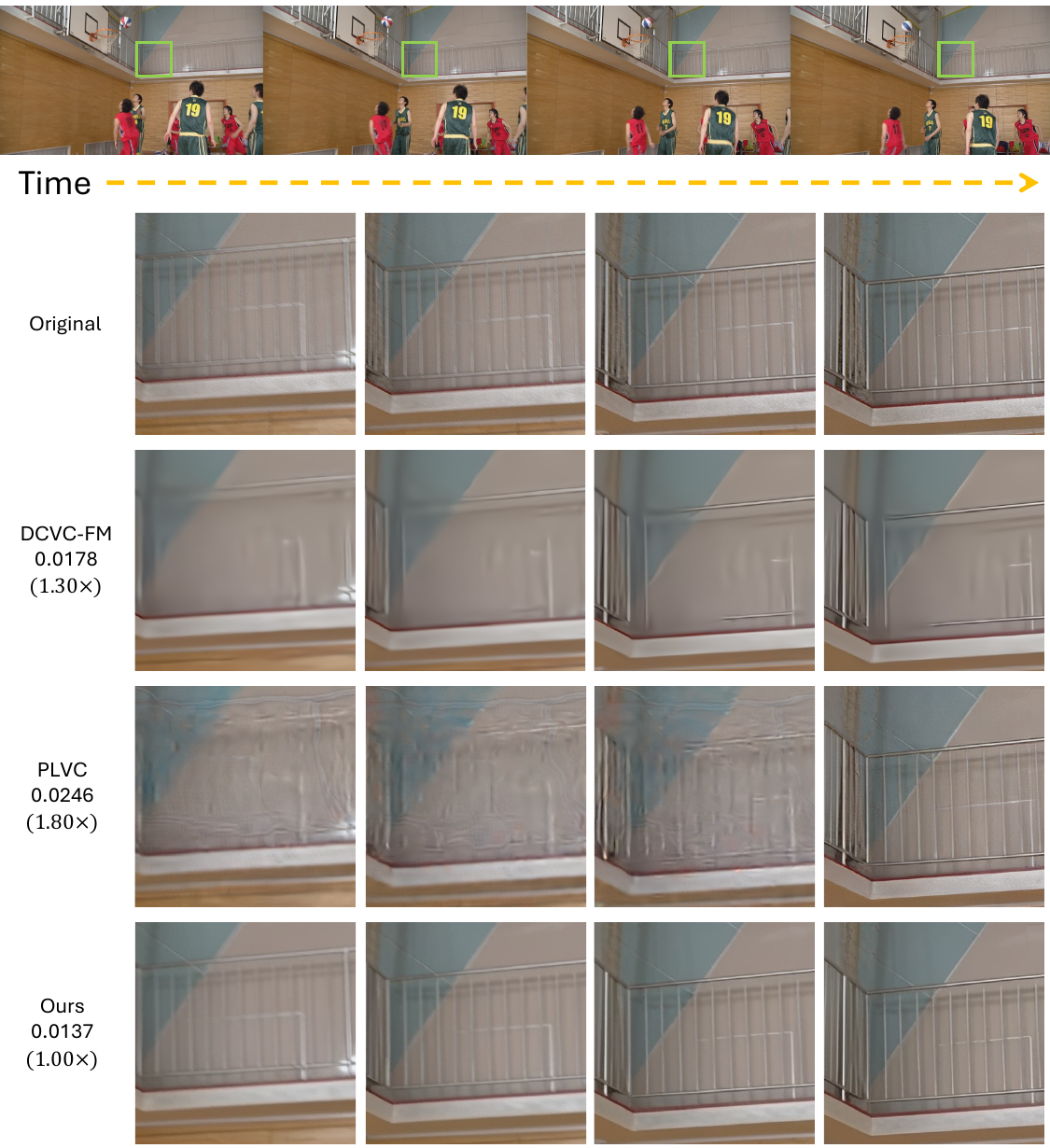}
    \vspace{-1mm}
    \caption{
        The proposed S$^2$VC accurately reconstructs the railing structure, whereas the neural codec DCVC-FM fails to recover its geometry, and the perceptual codec PLVC introduces additional temporally inconsistent noise.
    }
    \vspace{-2mm}
    \label{fig:supp_visual_2}
\end{figure*}

\begin{figure*}[t]
    \centering
    \includegraphics[width=1.0\linewidth]{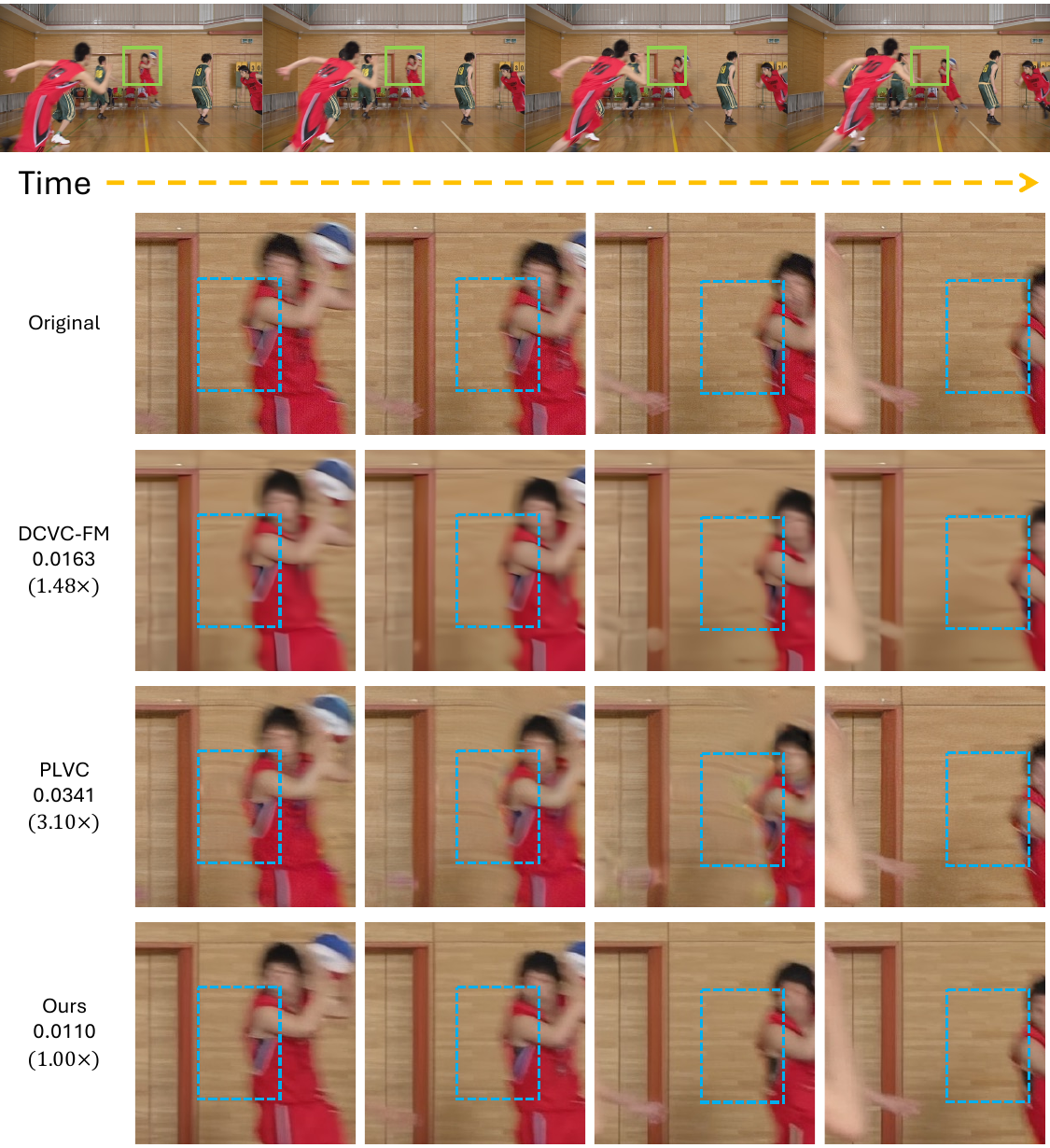}
    \vspace{-1mm}
    \caption{
        Additional visual examples in a motion scenario across neural codecs.
        The proposed S$^2$VC delivers the clearest and most temporally consistent texture on the wall, even as the subject moves over it.
        In contrast, DCVC-FM generates blurry details, while PLVC produces distorted textures that are inconsistent over time.
    }
    \vspace{-2mm}
    \label{fig:supp_visual_3}
\end{figure*}

\begin{figure*}[t]
    \centering
    \includegraphics[width=1.0\linewidth]{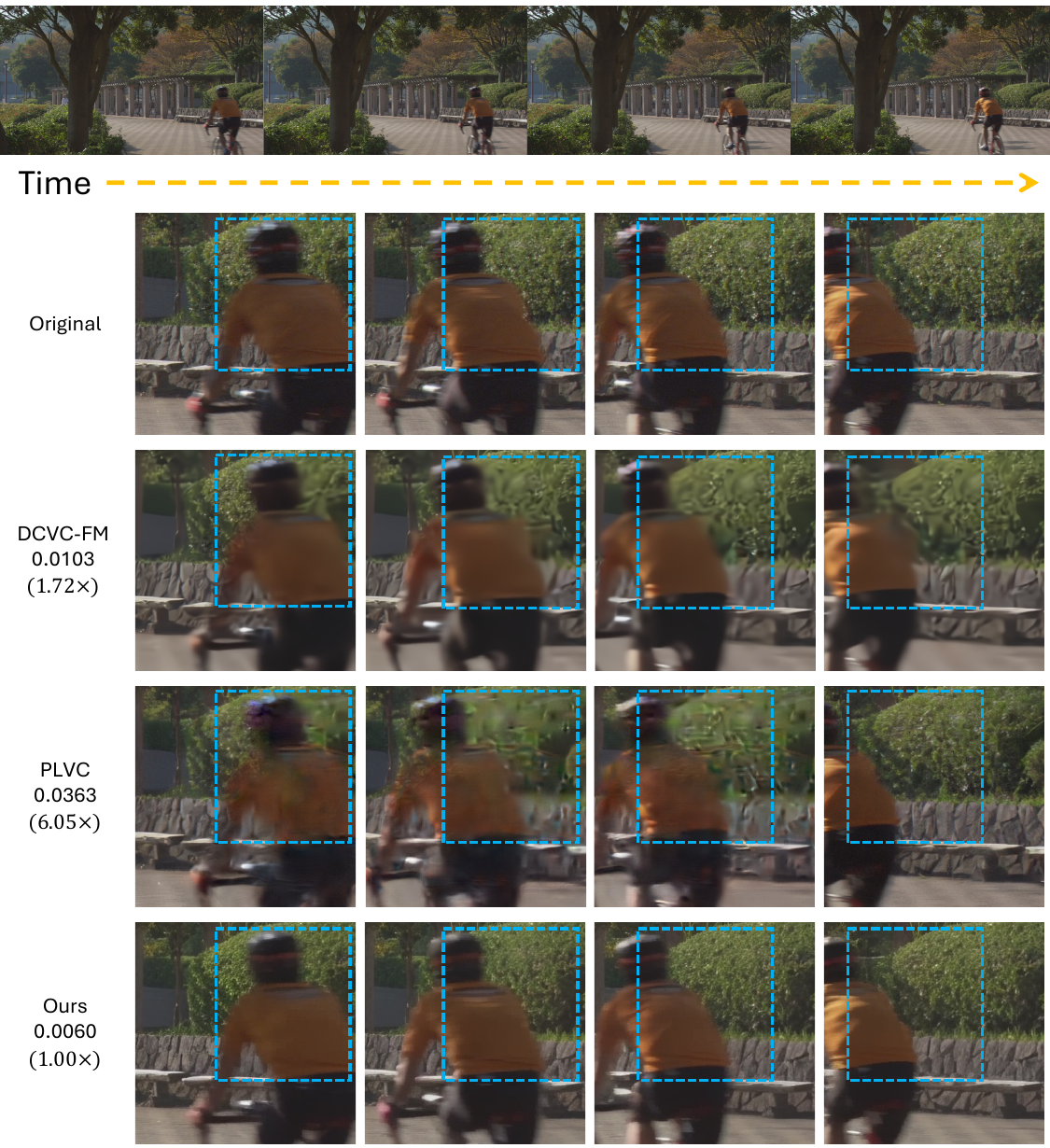}
    \vspace{-1mm}
    \caption{
        Additional visual examples in a motion scenario across neural codecs.
        The proposed S$^2$VC synthesizes consistent texture on the background trees, unaffected by the moving subject.
        In contrast, DCVC-FM generates blurry details, while PLVC produces jitter artifacts on both the subject and background, resulting in temporal inconsistencies.
    }
    \vspace{-2mm}
    \label{fig:supp_visual_4}
\end{figure*}

Temporal consistency comparisons across neural codecs are provided in Fig.~\ref{fig:supp_visual_2}, Fig.~\ref{fig:supp_visual_3} and Fig.~\ref{fig:supp_visual_4}.
S$^2$VC yields stable reconstructions for both moving objects and background regions.
In contrast, DCVC-FM produces blurry results, while PLVC introduces jitter and flicker artifacts in motion areas, leading to degraded temporal coherence.

These results demonstrate that S$^2$VC excels not only in generating detailed frame-level content but also in maintaining strong temporal consistency.

\subsection{Coding Latency}
We compare the coding latency of S$^2$VC with the neural codec DCVC-FM, evaluating both encoding and decoding speed in frames-per-second (fps).
All tests are conducted on an A100 GPU using 1920$\times$1080 video inputs.
As reported in Table~\ref{tab:supp_1}, the motion-vector-free design of S$^2$VC enables faster encoding than DCVC-FM, demonstrating its efficiency on the encoding side.
In contrast, the large one-step diffusion generator results in slower decoding, which is expected given the high generative capacity required for perceptual reconstruction.
As demonstrated by the RD results in the main paper, this computational cost translates into substantially improved perceptual quality.

Since the primary objective of this work is to investigate perceptual optimization with single-step diffusion in video coding, runtime optimization is not the main focus.
Future improvements, such as integerization of the compression module or distilling a smaller diffusion model, could further reduce coding latency.

\begin{table}[h]
\centering
\caption{Coding speed on 1920$\times$1080 video with an A100 GPU}
\vspace{-2mm}
\label{tab:supp_1}
\begin{tabular}{@{}lcc@{}}
\toprule
\multicolumn{1}{c}{Model} & Enc. fps & Dec. fps \\ \midrule
DCVC-FM~\cite{li2024neural} & 5.0 & 5.9 \\
S$^2$VC (Ours)& 6.6 & 1.27  \\ \bottomrule
\end{tabular}
\end{table}

\section{Model Training}
This section outlines the training protocol of the proposed S$^2$VC model, including the pretrained components, dataset, and overall training procedure.
We also describe the random-resolution strategy and the random group-wise cascade scheme used during training.

\begin{figure}[t]
    \centering
    \includegraphics[width=1.0\linewidth]{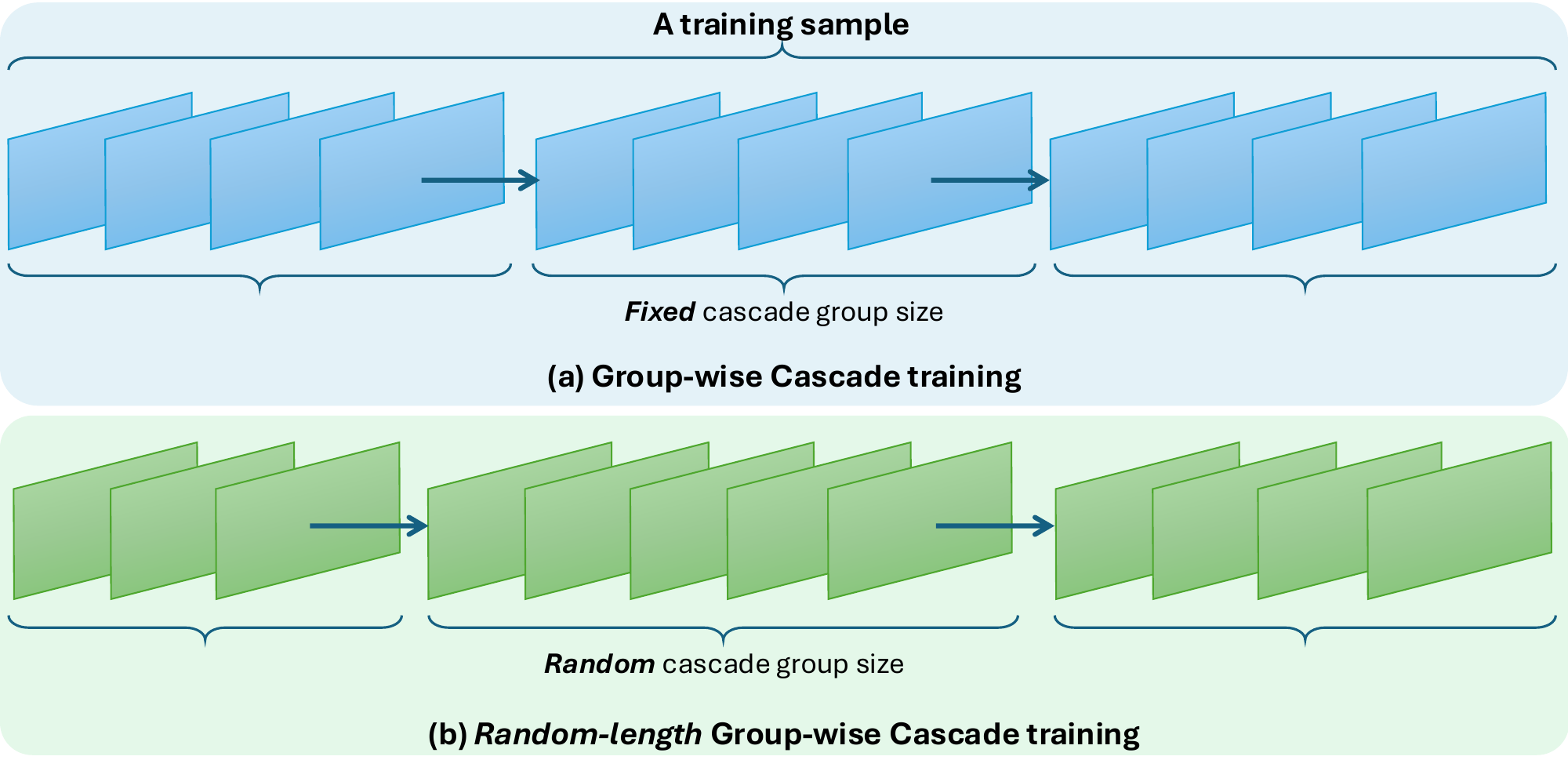}
    \vspace{-6mm}
    \caption{Comparison of two types of group-wise training. The proposed variant in \textbf{(b)} increases training diversity by randomly choosing the length of each group during cascade training.}
    \vspace{-4mm}
    \label{fig:supp_1}
\end{figure}

\vspace{1mm}
\noindent \textbf{Pretrained Models.}
To initialize the one-step diffusion model, we use the SD1.5 variant of DMD2~\cite{yin2024improved}, leveraging its pretrained generative prior to enhance training.
For semantic guidance, we adopt the \textit{dinov3-convnext-base} model~\cite{simeoni2025dinov3}, as its convolutional architecture efficiently supports variable spatial resolutions.
For the I-frame codec, we use the pretrained OneDC~\cite{xue2025one} models at five bitrate levels.

\vspace{1mm}
\noindent \textbf{Training Dataset.}
We train our model on the OpenVid-HD dataset~\cite{nan2024openvid}, a high-quality HD collection containing approximately 0.4M video clips.

\vspace{1mm}
\noindent \textbf{Training Procedure.}
The training loss is defined as the sum of rate, distortion, semantic, and motion losses:
\begin{equation}
    L = \frac{1}{T}\sum_{t=1}^{t=T}{(\lambda R + L_{\text{D}} + \alpha L_\text{sem} + \beta L_\text{motion})}
\end{equation}
where $T$ is the number of frames per iteration, and $R$ is the bitrate from the spatial-temporal entropy model~\cite{jia2025towards}.
The parameter $\lambda$ controls the rate-distortion trade-off, and we set $\lambda$ to $\{0.2, 0.4, 0.8, 1.2, 1.8, 2.6, 3.7, 5.2 \}$ for different bitrate.
Other terms are defined as follows:
\begin{align}
    L_\text{sem} &= ||E_{\text{DINO}}(x_t) - P_{\text{aux}}(s_{t})||_{1} \\
    L_{\text{D}} &= ||x_t-\hat{x}_t||_{1} + L_\text{LPIPS}(x_t, \hat{x}_t) \\
    L_\text{motion} &= ||O(x_{t-1}, x_{t}) - O(\hat{x}_{t-1}, \hat{x}_{t})||_{1}
\end{align}
where $O$ refers to the pretrained RAFT model~\cite{teed2020raft}, used to improve temporal consistency through optical flow alignment.
We set $\alpha=0.001$ and $\beta=0.25$ during training, and optimize using AdamW~\cite{loshchilov2017decoupled} with a total batch size of 16 across all GPUs.
We gradually increase the frame number T and decrease the learning rate to stabilize training:
\begin{itemize}
    \item For the first 20,000 iterations, we linearly increase T from 2 to 32, while linearly decreasing the learning rate from 1e-4 to 5e-6.
    \item For the next 10,000 iterations, we fix T=32 and set the learning rate to 1e-6.
\end{itemize}

\noindent All training tasks are conducted on 4$\times$A100 80GB GPUs, taking approximately 14 days.

\begin{table}[h]
\vspace{-2mm}
\centering
\caption{Parameter Count}
\vspace{-3mm}
\label{tab:supp_2}
\begin{tabular}{@{}ll@{}}
\toprule
Module & Parameters \\ \midrule
Codec Module & 144M \\
Diffusion Module & 1152M \\
\quad \textcolor{gray}{Base U-Net \MakeUppercase{*}} & \textcolor{gray}{860M} \\
\quad \textcolor{gray}{P-frame LoRA} & \textcolor{gray}{68M} \\
\quad \textcolor{gray}{TCG blocks} & \textcolor{gray}{224M} \\
VAE decoder &  50M \\ \midrule
\textbf{Total} & \textbf{1346M} \\ \bottomrule
\multicolumn{2}{l}{\small
\textit{\textcolor{gray}{Auxiliary predictor (Training only): 36M.}}
} \\
\multicolumn{2}{l}{\small
\textit{\textcolor{gray}{Module with \MakeUppercase{*} is frozen.}}
}
\end{tabular}
\vspace{-6mm}
\end{table}

\vspace{1mm}
\noindent \textbf{Random Resolution Training.}
To enable the one-step diffusion model to generalize across videos with varying resolutions, we train it using randomly selected resolutions from
\begin{equation}
    \{512\times512, 512\times768, 768\times1024, 1024\times1024\} \nonumber
\end{equation}
For each training iteration, a resolution is randomly sampled from this set, and a corresponding patch is cropped from a random spatial location within the training video.

\vspace{1mm}
\noindent \textbf{Random Group-wise Cascade Training.}
We adopt the group-wise cascade training scheme following ECVC~\cite{jiang2025ecvc} to reduce memory consumption when training the codec.
To increase training diversity and enhance generalization, we randomize the cascade propagation length with a maximum of 9 frames per iteration, as shown in Fig.~\ref{fig:supp_1}.

\section{Model Architecture}
This section first describes the LoRA~\cite{hu2022lora} configuration applied to the diffusion U-Net and provides the associated parameter count.
It then outlines the architecture of all modules used in the proposed model.

\vspace{1mm}
\noindent \textbf{LoRA Configuration and Parameter Count.}
We incorporate LoRA layers into all blocks of the diffusion U-Net (DMD2 distilled SD1.5 version~\cite{yin2024improved}) using the PEFT library~\cite{peft}.
Following the configuration of OneDC~\cite{xue2025one}, we set the LoRA rank to 64, the scaling factor (LoRA~$\alpha$) to 8.0, and the dropout rate to 0.0.
The detailed parameter statistics are provided in Table~\ref{tab:supp_2}.

\begin{figure}[h]
    \centering
    \includegraphics[width=1.0\linewidth]{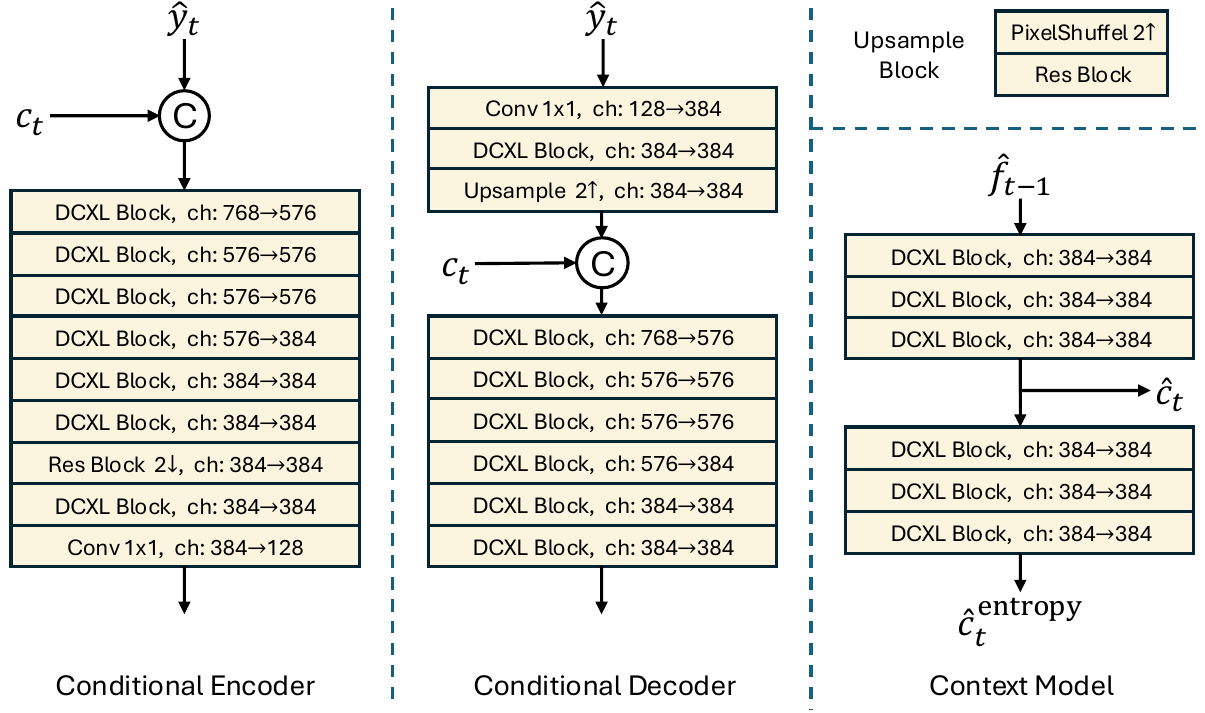}
    \vspace{-5mm}
    \caption{Architecture of the Conditional Encoder/Decoder and Context model used in S$^2$VC.}
    \vspace{-2mm}
    \label{fig:supp_2}
\end{figure}

\begin{figure}[h]
    \centering
    \includegraphics[width=1.0\linewidth]{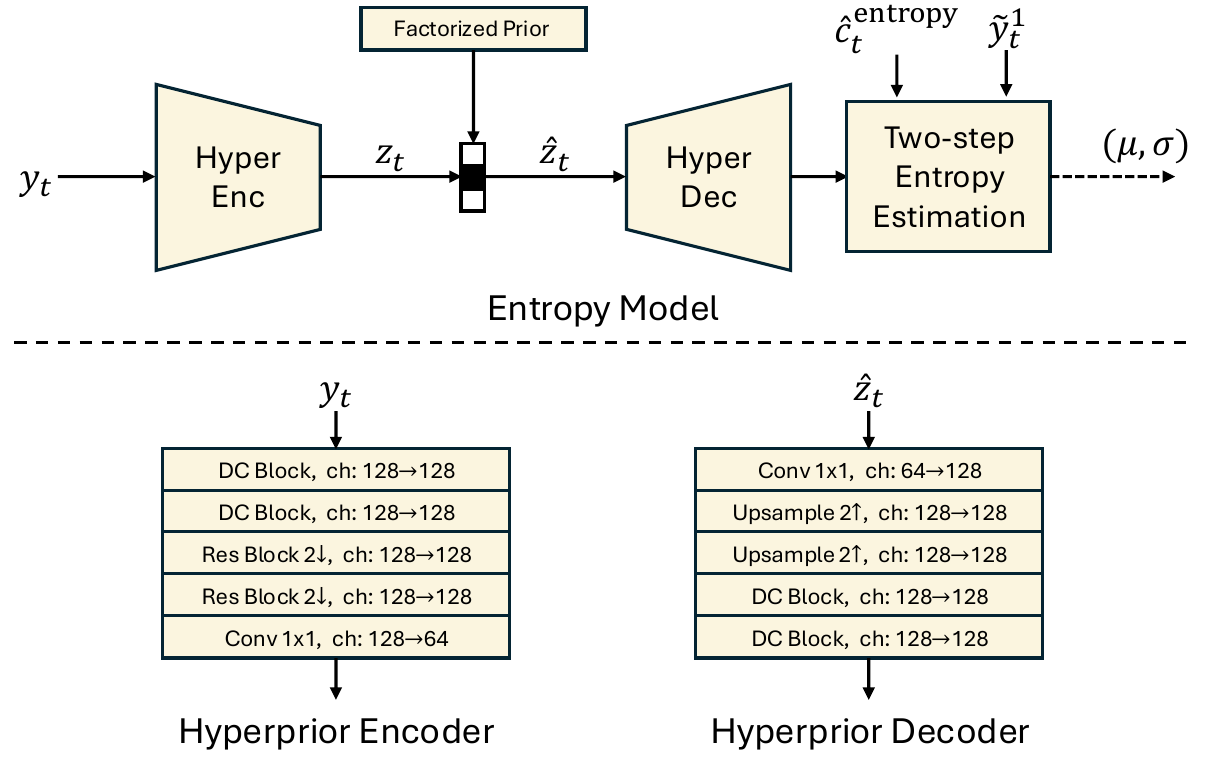}
    \vspace{-5mm}
    \caption{Architecture of the Entropy Model used in S$^2$VC. For Two-step Entropy Estimation, we use the implementation in~\cite{jia2025towards}.}
    \vspace{-2mm}
    \label{fig:supp_3}
\end{figure}

\begin{figure}[h]
    \centering
    \includegraphics[width=1.0\linewidth]{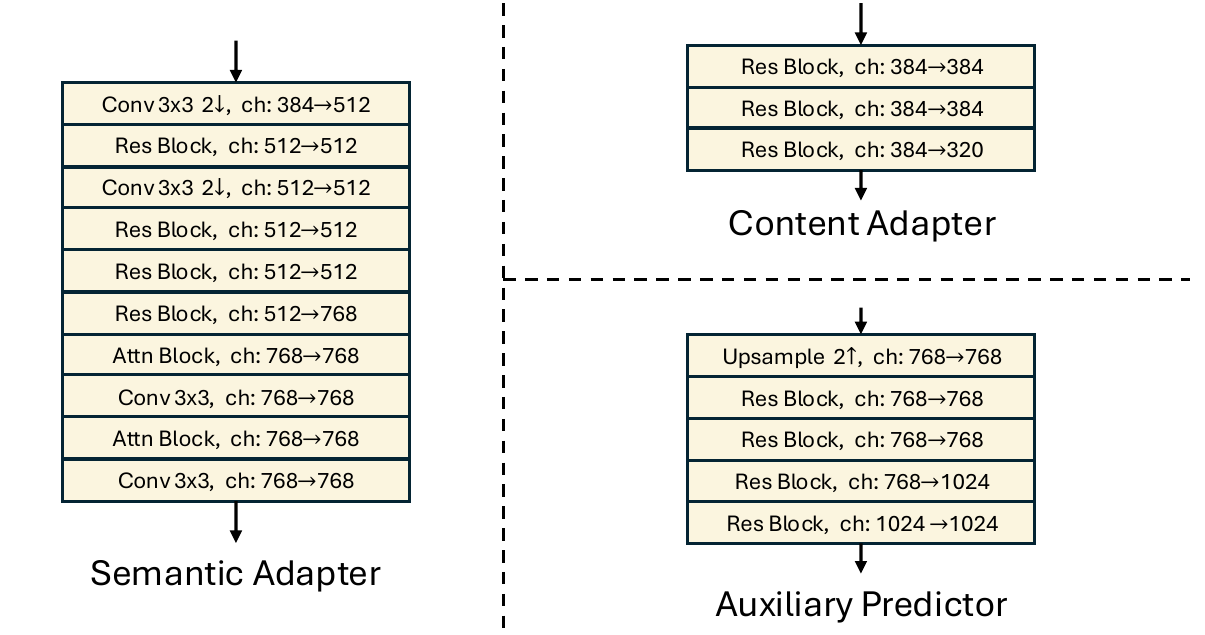}
    \vspace{-5mm}
    \caption{Architecture of the Content/Semantic Adapter and Auxiliary predictor.}
    \vspace{-2mm}
    \label{fig:supp_4}
\end{figure}

\begin{figure}[h]
    \centering
    \includegraphics[width=1.0\linewidth]{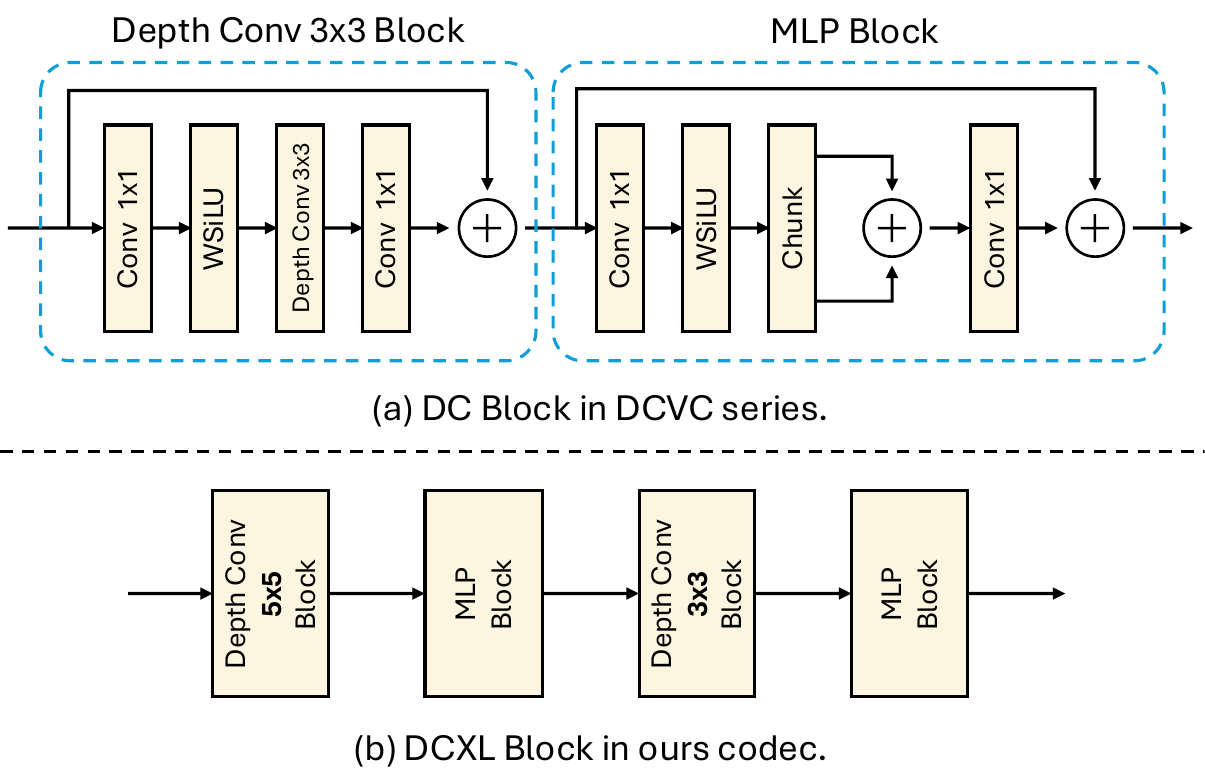}
    \vspace{-5mm}
    \caption{DC / DCXL blocks used in S$^2$VC. DC Block is from~\cite{jia2025towards}, and we enlarge it for higher model capacity.}
    \vspace{-2mm}
    \label{fig:supp_5}
\end{figure}

\begin{figure}[h]
    \centering
    \includegraphics[width=1.0\linewidth]{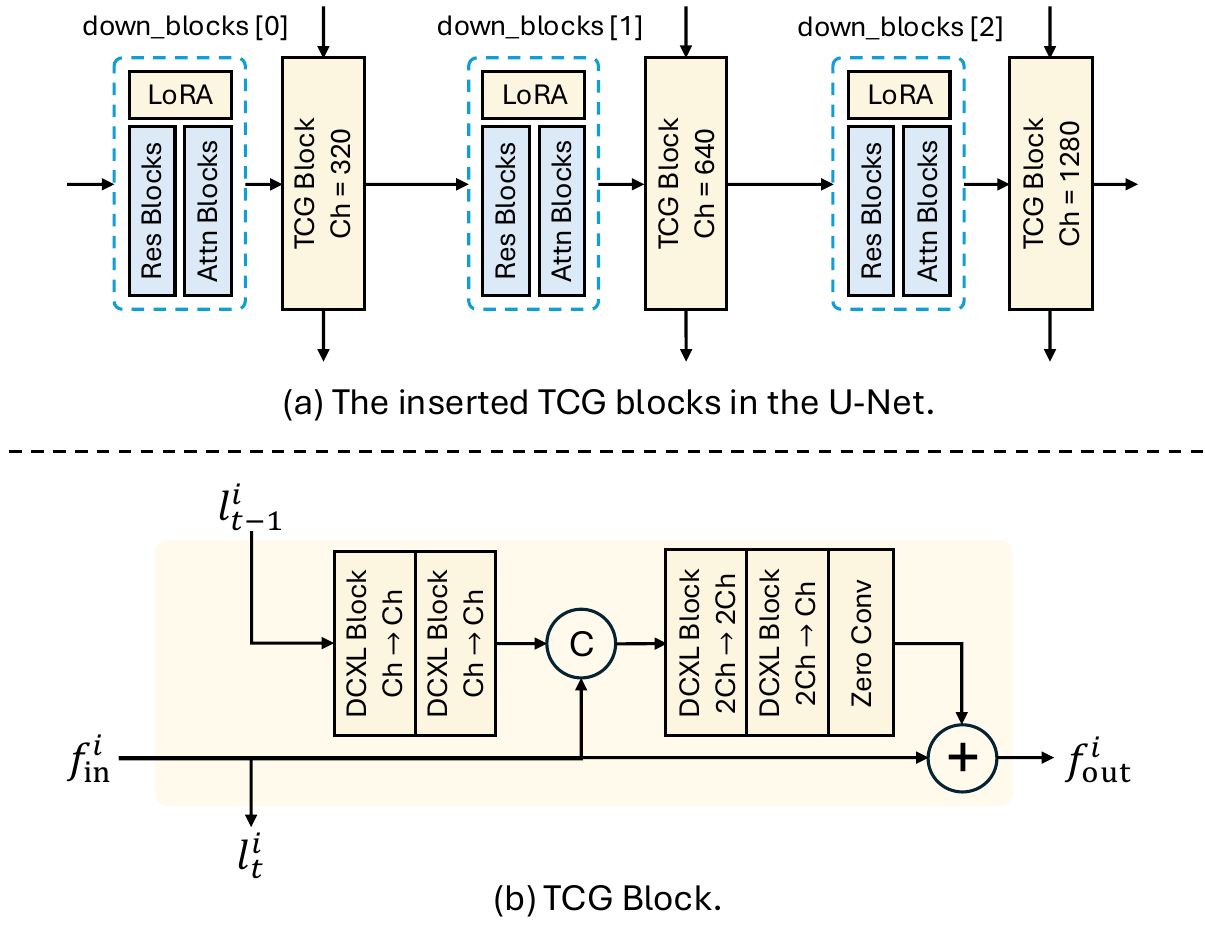}
    \vspace{-5mm}
    \caption{
        TCG block insertion in the diffusion U-Net.
        We build upon the \textit{UNet2DConditionModel} architecture from Diffusers~\cite{von-platen-etal-2022-diffusers} and insert TCG blocks into the first three scales, placing one after each \textit{down\_blocks} stage.
        $l^{i}_{t}$ denotes the intermediate feature at the $i$-th scale of frame $t$.
        $f^{i}_{\text{in}}$ / $f^{i}_{\text{out}}$ are input and output of a TCG block.
    }
    \vspace{-2mm}
    \label{fig:supp_6}
\end{figure}

\vspace{1mm}
\noindent \textbf{Detailed Model Architecture.}
The complete architectures of all modules are illustrated in Fig.~\ref{fig:supp_2}–\ref{fig:supp_6}.
The two-step entropy estimation module follows the design in~\cite{jia2025towards}, with channel dimensions adjusted to match our framework.
The ResBlock and AttnBlock structures are adopted from the Diffusers library~\cite{von-platen-etal-2022-diffusers}.

%% file: main.bib
@String(ICCV= {Int. Conf. Comput. Vis.})

@String(ICIP = {IEEE Int. Conf. Image Process.})

@String(ICLR = {Int. Conf. Learn. Represent.})

@String(IJCAI = {IJCAI})

@String(ICCV  = {ICCV})

@String(ICIP  = {ICIP})

@String(ICLR  = {ICLR})

@techreport{cisco2020internet,
  title        = {Cisco Annual Internet Report (2018–2023) White Paper},
  author       = {{Cisco Systems, Inc.}},
  year         = {2020},
  institution  = {Cisco Systems},
  url          = {https://www.cisco.com/c/en/us/solutions/collateral/executive-perspectives/annual-internet-report/white-paper-c11-741490.html},
}

@article{sullivan2012overview,
  title={Overview of the high efficiency video coding (HEVC) standard},
  author={Sullivan, Gary J and Ohm, Jens-Rainer and Han, Woo-Jin and Wiegand, Thomas},
  journal={IEEE Transactions on circuits and systems for video technology},
  volume={22},
  number={12},
  pages={1649--1668},
  year={2012},
  publisher={IEEE}
}

@article{bross2021overview,
  title={Overview of the versatile video coding (VVC) standard and its applications},
  author={Bross, Benjamin and Wang, Ye-Kui and Ye, Yan and Liu, Shan and Chen, Jianle and Sullivan, Gary J and Ohm, Jens-Rainer},
  journal={IEEE Transactions on Circuits and Systems for Video Technology},
  volume={31},
  number={10},
  pages={3736--3764},
  year={2021},
  publisher={IEEE}
}

@misc{jvet2025ecm,
  title={Explorations: Enhanced compression beyond VVC capability (ecm)},
  author={JVET},
  journal={MPEG Standards Exploration},
  year={2025},
  url={{https://www.mpeg.org/standards/Explorations/41}},
}

@article{balle2018variational,
  title={Variational image compression with a scale hyperprior},
  author={Ball{\'e}, Johannes and Minnen, David and Singh, Saurabh and Hwang, Sung Jin and Johnston, Nick},
  journal={arXiv preprint arXiv:1802.01436},
  year={2018}
}

@article{minnen2018joint,
  title={Joint autoregressive and hierarchical priors for learned image compression},
  author={Minnen, David and Ball{\'e}, Johannes and Toderici, George D},
  journal={Advances in neural information processing systems},
  volume={31},
  year={2018}
}

@article{wang2023evc,
  title={Evc: Towards real-time neural image compression with mask decay},
  author={Wang, Guo-Hua and Li, Jiahao and Li, Bin and Lu, Yan},
  journal={arXiv preprint arXiv:2302.05071},
  year={2023}
}

@inproceedings{lu2019dvc,
  title={Dvc: An end-to-end deep video compression framework},
  author={Lu, Guo and Ouyang, Wanli and Xu, Dong and Zhang, Xiaoyun and Cai, Chunlei and Gao, Zhiyong},
  booktitle={Proceedings of the IEEE/CVF conference on computer vision and pattern recognition},
  pages={11006--11015},
  year={2019}
}

@article{li2021deep,
  title={Deep contextual video compression},
  author={Li, Jiahao and Li, Bin and Lu, Yan},
  journal={Advances in Neural Information Processing Systems},
  volume={34},
  pages={18114--18125},
  year={2021}
}

@article{sheng2022temporal,
  title={Temporal context mining for learned video compression},
  author={Sheng, Xihua and Li, Jiahao and Li, Bin and Li, Li and Liu, Dong and Lu, Yan},
  journal={IEEE Transactions on Multimedia},
  volume={25},
  pages={7311--7322},
  year={2022},
  publisher={IEEE}
}

@inproceedings{li2023neural,
  title={Neural video compression with diverse contexts},
  author={Li, Jiahao and Li, Bin and Lu, Yan},
  booktitle={Proceedings of the IEEE/CVF conference on computer vision and pattern recognition},
  pages={22616--22626},
  year={2023}
}

@inproceedings{li2024neural,
  title={Neural video compression with feature modulation},
  author={Li, Jiahao and Li, Bin and Lu, Yan},
  booktitle={Proceedings of the IEEE/CVF Conference on Computer Vision and Pattern Recognition},
  pages={26099--26108},
  year={2024}
}

@inproceedings{jia2025towards,
  title={Towards practical real-time neural video compression},
  author={Jia, Zhaoyang and Li, Bin and Li, Jiahao and Xie, Wenxuan and Qi, Linfeng and Li, Houqiang and Lu, Yan},
  booktitle={Proceedings of the Computer Vision and Pattern Recognition Conference},
  pages={12543--12552},
  year={2025}
}

@inproceedings{jiang2025ecvc,
  title={Ecvc: Exploiting non-local correlations in multiple frames for contextual video compression},
  author={Jiang, Wei and Li, Junru and Zhang, Kai and Zhang, Li},
  booktitle={Proceedings of the Computer Vision and Pattern Recognition Conference},
  pages={7331--7341},
  year={2025}
}

@article{mentzer2020high,
  title={High-fidelity generative image compression},
  author={Mentzer, Fabian and Toderici, George D and Tschannen, Michael and Agustsson, Eirikur},
  journal={Advances in neural information processing systems},
  volume={33},
  pages={11913--11924},
  year={2020}
}

@inproceedings{muckley2023improving,
  title={Improving statistical fidelity for neural image compression with implicit local likelihood models},
  author={Muckley, Matthew J and El-Nouby, Alaaeldin and Ullrich, Karen and J{\'e}gou, Herv{\'e} and Verbeek, Jakob},
  booktitle={International Conference on Machine Learning},
  pages={25426--25443},
  year={2023},
  organization={PMLR}
}

@inproceedings{careil2023towards,
  title={Towards image compression with perfect realism at ultra-low bitrates},
  author={Careil, Marlene and Muckley, Matthew J and Verbeek, Jakob and Lathuili{\`e}re, St{\'e}phane},
  booktitle={The Twelfth International Conference on Learning Representations},
  year={2023}
}

@article{li2024towards,
  title={Towards Extreme Image Compression with Latent Feature Guidance and Diffusion Prior},
  author={Li, Zhiyuan and Zhou, Yanhui and Wei, Hao and Ge, Chenyang and Jiang, Jingwen},
  journal={IEEE Transactions on Circuits and Systems for Video Technology},
  year={2024},
  publisher={IEEE}
}

@article{zhang2025stablecodec,
  title={StableCodec: Taming One-Step Diffusion for Extreme Image Compression},
  author={Zhang, Tianyu and Luo, Xin and Li, Li and Liu, Dong},
  journal={arXiv preprint arXiv:2506.21977},
  year={2025}
}

@article{xue2025one,
  title={One-Step Diffusion-Based Image Compression with Semantic Distillation},
  author={Xue, Naifu and Jia, Zhaoyang and Li, Jiahao and Li, Bin and Zhang, Yuan and Lu, Yan},
  journal={arXiv preprint arXiv:2505.16687},
  year={2025}
}

@inproceedings{zhang2021dvc,
  title={DVC-P: Deep video compression with perceptual optimizations},
  author={Zhang, Saiping and Mrak, Marta and Herranz, Luis and Blanch, Marc G{\'o}rriz and Wan, Shuai and Yang, Fuzheng},
  booktitle={2021 International Conference on Visual Communications and Image Processing (VCIP)},
  pages={1--5},
  year={2021},
  organization={IEEE}
}

@inproceedings{mentzer2022neural,
  title={Neural video compression using gans for detail synthesis and propagation},
  author={Mentzer, Fabian and Agustsson, Eirikur and Ball{\'e}, Johannes and Minnen, David and Johnston, Nick and Toderici, George},
  booktitle={European Conference on Computer Vision},
  pages={562--578},
  year={2022},
  organization={Springer}
}

@inproceedings{li2023high,
  title={High visual-fidelity learned video compression},
  author={Li, Meng and Shi, Yibo and Wang, Jing and Huang, Yunqi},
  booktitle={Proceedings of the 31st ACM International Conference on Multimedia},
  pages={8057--8066},
  year={2023}
}

@inproceedings{yang2022perceptual,
  title={Perceptual Learned Video Compression with Recurrent Conditional GAN.},
  author={Yang, Ren and Timofte, Radu and Van Gool, Luc},
  booktitle={IJCAI},
  pages={1537--1544},
  year={2022}
}

@article{liu20242,
  title={I\textsuperscript{2}VC: A Unified Framework for Intra-\& Inter-frame Video Compression},
  author={Liu, Meiqin and Xu, Chenming and Gu, Yukai and Yao, Chao and Zhao, Yao},
  journal={arXiv preprint arXiv:2405.14336},
  year={2024}
}

@article{ma2025diffusion,
  title={Diffusion-based perceptual neural video compression with temporal diffusion information reuse},
  author={Ma, Wenzhuo and Chen, Zhenzhong},
  journal={arXiv preprint arXiv:2501.13528},
  year={2025}
}

@article{ma2025diffvc,
  title={DiffVC-OSD: One-Step Diffusion-based Perceptual Neural Video Compression Framework},
  author={Ma, Wenzhuo and Chen, Zhenzhong},
  journal={arXiv preprint arXiv:2508.07682},
  year={2025}
}

@misc{rombach2021highresolution,
      title={High-Resolution Image Synthesis with Latent Diffusion Models}, 
      author={Robin Rombach and Andreas Blattmann and Dominik Lorenz and Patrick Esser and Björn Ommer},
      year={2021},
      eprint={2112.10752},
      archivePrefix={arXiv},
      primaryClass={cs.CV}
}

@inproceedings{song2023consistency,
  title={Consistency Models},
  author={Song, Yang and Dhariwal, Prafulla and Chen, Mark and Sutskever, Ilya},
  booktitle={International Conference on Machine Learning},
  pages={32211--32252},
  year={2023},
  organization={PMLR}
}

@inproceedings{yin2024one,
  title={One-step diffusion with distribution matching distillation},
  author={Yin, Tianwei and Gharbi, Micha{\"e}l and Zhang, Richard and Shechtman, Eli and Durand, Fredo and Freeman, William T and Park, Taesung},
  booktitle={Proceedings of the IEEE/CVF conference on computer vision and pattern recognition},
  pages={6613--6623},
  year={2024}
}

@article{yin2024improved,
  title={Improved distribution matching distillation for fast image synthesis},
  author={Yin, Tianwei and Gharbi, Micha{\"e}l and Park, Taesung and Zhang, Richard and Shechtman, Eli and Durand, Fredo and Freeman, Bill},
  journal={Advances in neural information processing systems},
  volume={37},
  pages={47455--47487},
  year={2024}
}

@article{liu2025ultravsr,
  title={UltraVSR: Achieving Ultra-Realistic Video Super-Resolution with Efficient One-Step Diffusion Space},
  author={Liu, Yong and Pan, Jinshan and Li, Yinchuan and Dong, Qingji and Zhu, Chao and Guo, Yu and Wang, Fei},
  journal={arXiv preprint arXiv:2505.19958},
  year={2025}
}

@article{sun2025one,
  title={One-Step Diffusion for Detail-Rich and Temporally Consistent Video Super-Resolution},
  author={Sun, Yujing and Sun, Lingchen and Liu, Shuaizheng and Wu, Rongyuan and Zhang, Zhengqiang and Zhang, Lei},
  journal={arXiv preprint arXiv:2506.15591},
  year={2025}
}

@article{guo2023animatediff,
  title={Animatediff: Animate your personalized text-to-image diffusion models without specific tuning},
  author={Guo, Yuwei and Yang, Ceyuan and Rao, Anyi and Liang, Zhengyang and Wang, Yaohui and Qiao, Yu and Agrawala, Maneesh and Lin, Dahua and Dai, Bo},
  journal={arXiv preprint arXiv:2307.04725},
  year={2023}
}

@article{blattmann2023stable,
  title={Stable video diffusion: Scaling latent video diffusion models to large datasets},
  author={Blattmann, Andreas and Dockhorn, Tim and Kulal, Sumith and Mendelevitch, Daniel and Kilian, Maciej and Lorenz, Dominik and Levi, Yam and English, Zion and Voleti, Vikram and Letts, Adam and others},
  journal={arXiv preprint arXiv:2311.15127},
  year={2023}
}

@article{simeoni2025dinov3,
  title={Dinov3},
  author={Sim{\'e}oni, Oriane and Vo, Huy V and Seitzer, Maximilian and Baldassarre, Federico and Oquab, Maxime and Jose, Cijo and Khalidov, Vasil and Szafraniec, Marc and Yi, Seungeun and Ramamonjisoa, Micha{\"e}l and others},
  journal={arXiv preprint arXiv:2508.10104},
  year={2025}
}

@article{leng2025repa,
  title={Repa-e: Unlocking vae for end-to-end tuning with latent diffusion transformers},
  author={Leng, Xingjian and Singh, Jaskirat and Hou, Yunzhong and Xing, Zhenchang and Xie, Saining and Zheng, Liang},
  journal={arXiv preprint arXiv:2504.10483},
  year={2025}
}

@inproceedings{teed2020raft,
  title={Raft: Recurrent all-pairs field transforms for optical flow},
  author={Teed, Zachary and Deng, Jia},
  booktitle={European conference on computer vision},
  pages={402--419},
  year={2020},
  organization={Springer}
}

@article{loshchilov2017decoupled,
  title={Decoupled weight decay regularization},
  author={Loshchilov, Ilya and Hutter, Frank},
  journal={arXiv preprint arXiv:1711.05101},
  year={2017}
}

@article{hu2022lora,
  title={Lora: Low-rank adaptation of large language models.},
  author={Hu, Edward J and Shen, Yelong and Wallis, Phillip and Allen-Zhu, Zeyuan and Li, Yuanzhi and Wang, Shean and Wang, Lu and Chen, Weizhu and others},
  journal={ICLR},
  volume={1},
  number={2},
  pages={3},
  year={2022}
}

@article{bjontegaard2001calculation,
  title={Calculation of average PSNR differences between RD-curves},
  author={Bjontegaard, Gisle},
  journal={ITU SG16 Doc. VCEG-M33},
  year={2001}
}

@inproceedings{wang2003multiscale,
  author={Wang, Z. and Simoncelli, E.P. and Bovik, A.C.},
  booktitle={The Thrity-Seventh Asilomar Conference on Signals, Systems \& Computers, 2003}, 
  title={Multiscale structural similarity for image quality assessment}, 
  year={2003},
  volume={2},
  number={},
  pages={1398-1402 Vol.2},
  doi={10.1109/ACSSC.2003.1292216}
}

@article{nan2024openvid,
  title={Openvid-1m: A large-scale high-quality dataset for text-to-video generation},
  author={Nan, Kepan and Xie, Rui and Zhou, Penghao and Fan, Tiehan and Yang, Zhenheng and Chen, Zhijie and Li, Xiang and Yang, Jian and Tai, Ying},
  journal={arXiv preprint arXiv:2407.02371},
  year={2024}
}

@inproceedings{mercat2020uvg,
  title={UVG dataset: 50/120fps 4K sequences for video codec analysis and development},
  author={Mercat, Alexandre and Viitanen, Marko and Vanne, Jarno},
  booktitle={Proceedings of the 11th ACM multimedia systems conference},
  pages={297--302},
  year={2020}
}

@inproceedings{wang2016mcl,
  title={MCL-JCV: a JND-based H. 264/AVC video quality assessment dataset},
  author={Wang, Haiqiang and Gan, Weihao and Hu, Sudeng and Lin, Joe Yuchieh and Jin, Lina and Song, Longguang and Wang, Ping and Katsavounidis, Ioannis and Aaron, Anne and Kuo, C-C Jay},
  booktitle={2016 IEEE international conference on image processing (ICIP)},
  pages={1509--1513},
  year={2016},
  organization={IEEE}
}

@inproceedings{johnson2016perceptual,
  title={Perceptual losses for real-time style transfer and super-resolution},
  author={Johnson, Justin and Alahi, Alexandre and Fei-Fei, Li},
  booktitle={Computer Vision--ECCV 2016: 14th European Conference, Amsterdam, The Netherlands, October 11-14, 2016, Proceedings, Part II 14},
  pages={694--711},
  year={2016},
  organization={Springer}
}

@article{ding2020image,
  title={Image quality assessment: Unifying structure and texture similarity},
  author={Ding, Keyan and Ma, Kede and Wang, Shiqi and Simoncelli, Eero P},
  journal={IEEE transactions on pattern analysis and machine intelligence},
  volume={44},
  number={5},
  pages={2567--2581},
  year={2020},
  publisher={IEEE}
}

@inproceedings{danier2022flolpips,
  title={FloLPIPS: A bespoke video quality metric for frame interpolation},
  author={Danier, Duolikun and Zhang, Fan and Bull, David},
  booktitle={2022 Picture Coding Symposium (PCS)},
  pages={283--287},
  year={2022},
  organization={IEEE}
}

@article{heusel2017gans,
  title={Gans trained by a two time-scale update rule converge to a local nash equilibrium},
  author={Heusel, Martin and Ramsauer, Hubert and Unterthiner, Thomas and Nessler, Bernhard and Hochreiter, Sepp},
  journal={Advances in neural information processing systems},
  volume={30},
  year={2017}
}

@inproceedings{DCVC-HEM,
  title={Hybrid Spatial-Temporal Entropy Modelling for Neural Video Compression},
  author={Li, Jiahao and Li, Bin and Lu, Yan},
  booktitle={Proceedings of the 30th ACM International Conference on Multimedia},
  pages={1503--1511},
  year={2022}
}

@InProceedings{xue2025dlf,
  author={Xue, Naifu and Jia, Zhaoyang and Li, Jiahao and Li, Bin and Zhang, Yuan and Lu, Yan},
  title={DLF: Extreme Image Compression with Dual-generative Latent Fusion},
  booktitle={Proceedings of the IEEE/CVF International Conference on Computer Vision (ICCV)},
  month = {Oct},
  year={2025},
}

@inproceedings{qi2024long,
  title={Long-term temporal context gathering for neural video compression},
  author={Qi, Linfeng and Jia, Zhaoyang and Li, Jiahao and Li, Bin and Li, Houqiang and Lu, Yan},
  booktitle={European Conference on Computer Vision},
  pages={305--322},
  year={2024},
  organization={Springer}
}

@inproceedings{jia2024generative,
  title={Generative latent coding for ultra-low bitrate image compression},
  author={Jia, Zhaoyang and Li, Jiahao and Li, Bin and Li, Houqiang and Lu, Yan},
  booktitle={Proceedings of the IEEE/CVF Conference on Computer Vision and Pattern Recognition},
  pages={26088--26098},
  year={2024}
}

@Misc{peft,
  title =        {{PEFT}: State-of-the-art Parameter-Efficient Fine-Tuning methods},
  author =       {Sourab Mangrulkar and Sylvain Gugger and Lysandre Debut and Younes Belkada and Sayak Paul and Benjamin Bossan},
  howpublished = {\url{https://github.com/huggingface/peft}},
  year =         {2022}
}

@misc{von-platen-etal-2022-diffusers,
  author = {Patrick von Platen and Suraj Patil and Anton Lozhkov and Pedro Cuenca and Nathan Lambert and Kashif Rasul and Mishig Davaadorj and Dhruv Nair and Sayak Paul and William Berman and Yiyi Xu and Steven Liu and Thomas Wolf},
  title = {Diffusers: State-of-the-art diffusion models},
  year = {2022},
  publisher = {GitHub},
  journal = {GitHub repository},
  howpublished = {\url{https://github.com/huggingface/diffusers}}
}

@article{qi2025generative,
  title={Generative latent coding for ultra-low bitrate image and video compression},
  author={Qi, Linfeng and Jia, Zhaoyang and Li, Jiahao and Li, Bin and Li, Houqiang and Lu, Yan},
  journal={IEEE Transactions on Circuits and Systems for Video Technology},
  year={2025},
  publisher={IEEE}
}

@misc{zheng2026generative,
      title={Generative Video Compression with One-Dimensional Latent Representation}, 
      author={Zihan Zheng and Zhaoyang Jia and Naifu Xue and Jiahao Li and Bin Li and Zongyu Guo and Xiaoyi Zhang and Zhenghao Chen and Houqiang Li and Yan Lu},
      year={2026},
      eprint={2603.15302},
      archivePrefix={arXiv},
      primaryClass={cs.CV},
      url={https://arxiv.org/abs/2603.15302}, 
}
